\documentclass{article} 
\usepackage{iclr2021_conference,times}

\usepackage{amsmath,amsfonts,bm}

\def\Figref#1{Figure~\ref{#1}}

\def\secref#1{section~\ref{#1}}

\def\eqref#1{equation~\ref{#1}}

\def\algref#1{algorithm~\ref{#1}}
\def\Algref#1{Algorithm~\ref{#1}}

\def\1{\bm{1}}

\def\rx{{\textnormal{x}}}

\def\rvx{{\mathbf{x}}}
\def\rvy{{\mathbf{y}}}

\def\mI{{\bm{I}}}

\DeclareMathAlphabet{\mathsfit}{\encodingdefault}{\sfdefault}{m}{sl}
\SetMathAlphabet{\mathsfit}{bold}{\encodingdefault}{\sfdefault}{bx}{n}

\newcommand{\pdata}{p_{\rm{data}}}

\newcommand{\E}{\mathbb{E}}

\newcommand{\KL}{D_{\mathrm{KL}}}
\newcommand{\Var}{\mathrm{Var}}

\usepackage{hyperref}  
\usepackage{url}
\usepackage{graphicx}
\usepackage{amsmath}
\usepackage{amssymb}
\usepackage{caption}
\usepackage{booktabs}
\usepackage{multirow}
\usepackage{mathtools}
\usepackage{adjustbox}
\usepackage{algorithm}
\usepackage{algorithmic}
\usepackage{wrapfig}
\title{Learning Energy-Based Models by Diffusion Recovery Likelihood}

\author{
 Ruiqi Gao \\
 UCLA\\
 \texttt{ruiqigao@ucla.edu} \\
 \And
 Yang Song \\
 Stanford University \\
 \texttt{yangsong@cs.stanford.edu} \\
 \And
 Ben Poole \\
 Google Brain \\
 \texttt{pooleb@google.com} \\
 \AND
 Ying Nian Wu\\ 
 UCLA \\
\texttt{ywu@stat.ucla.edu} \\
 \And
  Diederik P. Kingma\\
 Google Brain \\
 \texttt{durk@google.com}
}

\def\N{{\mathcal N}}
\def\trvx{\tilde{\rvx}}
\def\beps{\bm{\epsilon}}

\iclrfinalcopy 
\begin{document}

\maketitle

\begin{abstract}
While energy-based models (EBMs) exhibit a number of desirable properties, training and sampling on high-dimensional datasets remains challenging. Inspired by recent progress on diffusion probabilistic models, we present a diffusion recovery likelihood method to tractably learn and sample from a sequence of EBMs trained on increasingly noisy versions of a dataset. Each EBM is trained with recovery likelihood,  which maximizes the conditional probability of the data at a certain noise level given their noisy versions at a higher noise level. Optimizing recovery likelihood is more tractable than marginal likelihood, as sampling from the conditional distributions is much easier than sampling from the marginal distributions. After training, synthesized images can be generated by the sampling process that initializes from Gaussian white noise distribution and progressively samples the conditional distributions at decreasingly lower noise levels.  Our method generates high fidelity samples on various image datasets. On unconditional CIFAR-10 our method achieves FID 9.58 and inception score 8.30, superior to the majority of GANs. Moreover, we demonstrate that unlike previous work on EBMs, our long-run MCMC samples from the conditional distributions do not diverge and still represent realistic images, allowing us to accurately estimate the normalized density of data even for high-dimensional datasets. Our implementation is available at \url{https://github.com/ruiqigao/recovery_likelihood}.
\end{abstract}

\section{Introduction}

EBMs~\citep{lecun2006tutorial,ngiam2011learning,kim2016deep,zhao2016energy,goyal2017variational,xie2016theory,finn2016connection,gao2018learning,kumar2019maximum,nijkamp2019learning,du2019implicit,grathwohl2019your,desjardins2011tracking,gao2020flow,che2020your,grathwohl2020cutting,qiu2019unbiased,rhodes2020telescoping} are an appealing class of probabilistic models, which can be viewed as generative versions of  discriminators~\citep{jin2017introspective, lazarow2017introspective, lee2018wasserstein, grathwohl2020cutting}, yet can be learned from unlabeled data. Despite a number of desirable properties, two challenges remain for training EBMs on high-dimensional datasets. First, learning EBMs by maximum likelihood requires Markov Chain Monte Carlo (MCMC) to generate samples from the model, which can be extremely expensive. Second, as pointed out in ~\citet{nijkamp2019anatomy}, the energy potentials learned with non-convergent MCMC do not have a valid steady-state, in the sense that samples from long-run Markov chains can differ greatly from observed samples, making it difficult to evaluate the learned energy potentials.

Another line of work, originating from \citet{sohl2015deep}, is to learn from a diffused version of the data, which are obtained from the original data via a diffusion process that sequentially adds Gaussian white noise. From such diffusion data, one can learn the conditional model of the data at a certain noise level given their noisy versions at the higher noise level of the diffusion process. After learning the sequence of conditional models that invert the diffusion process, one can then generate synthesized images from Gaussian white noise images by ancestral sampling.  Building on \citet{sohl2015deep}, \citet{ho2020denoising} further developed the method, obtaining strong image synthesis results. 

Inspired by \citet{sohl2015deep} and \citet{ho2020denoising}, we propose a {\em diffusion recovery likelihood} method to tackle the challenge of training EBMs directly on a dataset by instead learning a sequence of EBMs for the {\em marginal} distributions of the diffusion process. The sequence of marginal EBMs are learned with recovery likelihoods that are defined as the conditional distributions that invert the diffusion process. Compared to  standard maximum likelihood estimation (MLE) of EBMs, learning marginal EBMs by diffusion recovery likelihood only requires sampling from the conditional distributions, which is much easier than sampling from the marginal distributions. After learning the marginal EBMs, we can generate synthesized images by a sequence of conditional samples initialized from the Gaussian white noise distribution. Unlike \citet{ho2020denoising} that approximates the reverse process by normal distributions, in our case the conditional distributions are derived from the marginal EBMs, which are more flexible. The framework of recovery likelihood was originally proposed in~\citet{bengio2013generalized}. In our work, we adapt it to learning the sequence of marginal EBMs from the diffusion data.

Our work is also related to the denoising score matching method of ~\citet{vincent2011connection}, which was further developed by~\citet{song2019generative,song2020improved} for learning from diffusion data. The training objective used for diffusion probabilisitic models is a weighted version of the denoising score matching objective, as revealed by \citet{ho2020denoising}. These methods learn the score functions (the gradients of the energy functions) directly, instead of using the gradients of learned energy functions as in EBMs. On the other hand, \cite{saremi2018deep} parametrizes the score function as the gradient of a MLP energy function, and \cite{saremi2019neural} further unifies denoising score matching and neural empirical Bayes.

We demonstrate the efficacy of diffusion recovery likelihood on CIFAR-10, CelebA and LSUN datasets. The generated samples are of high fidelity and comparable to GAN-based methods. On CIFAR-10, we achieve FID 9.58 and inception score 8.30, exceeding existing methods of learning explicit EBMs to a large extent. We also demonstrate that diffusion recovery likelihood outperforms denoising score matching from diffusion data if we naively take the gradients of explicit energy functions as the score functions.  More interestingly, by using a thousand diffusion time steps, we demonstrate that even very long MCMC chains from the sequence of conditional distributions produce samples that represent realistic images. With the faithful long-run MCMC samples from the conditional distributions, we can accurately estimate the marginal partition function at zero noise level by importance sampling, and thus evaluate the normalized density of data under the EBM.

\begin{figure}[t]
\begin{center}
\includegraphics[width=.325\textwidth]{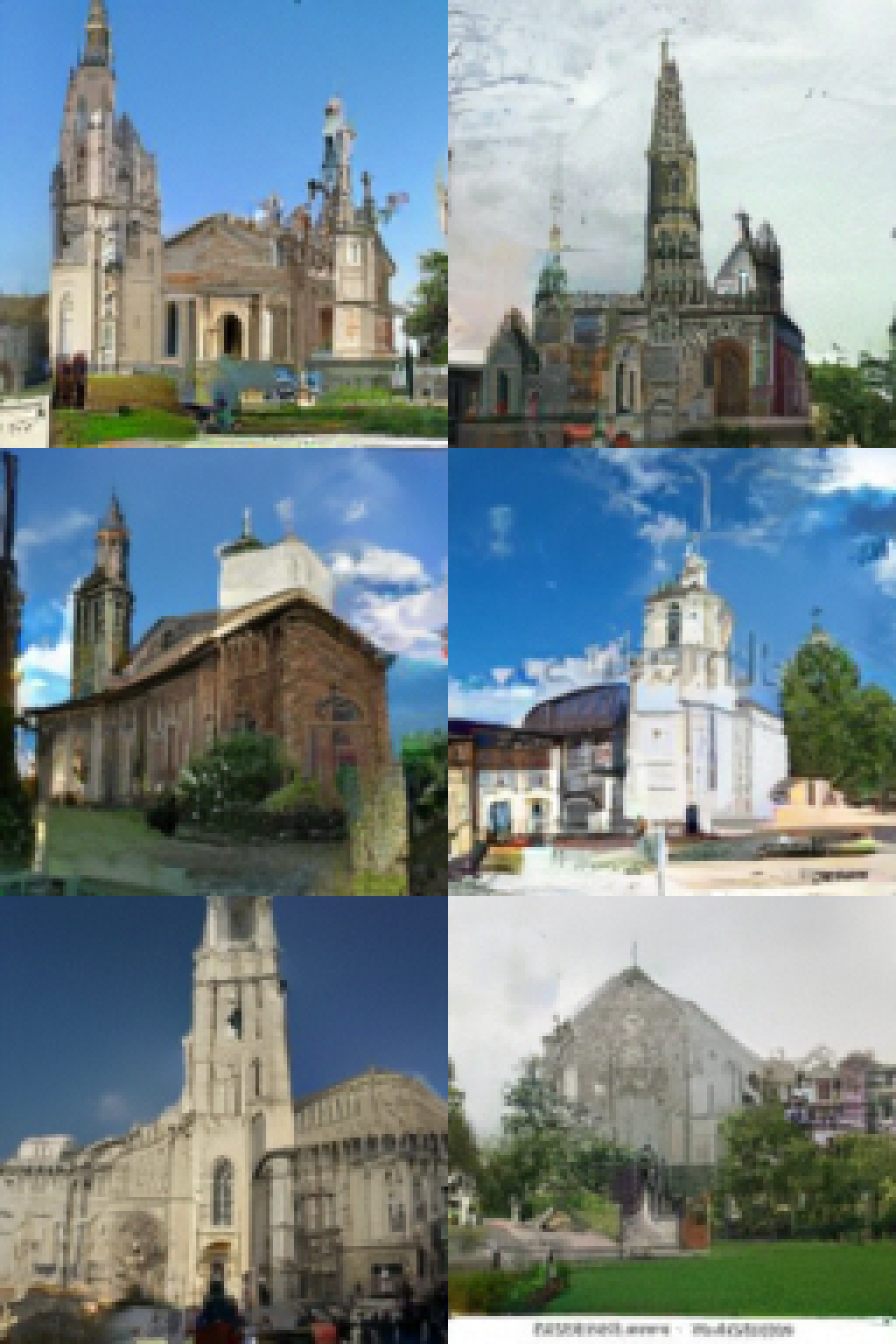} 
\includegraphics[width=.325\textwidth]{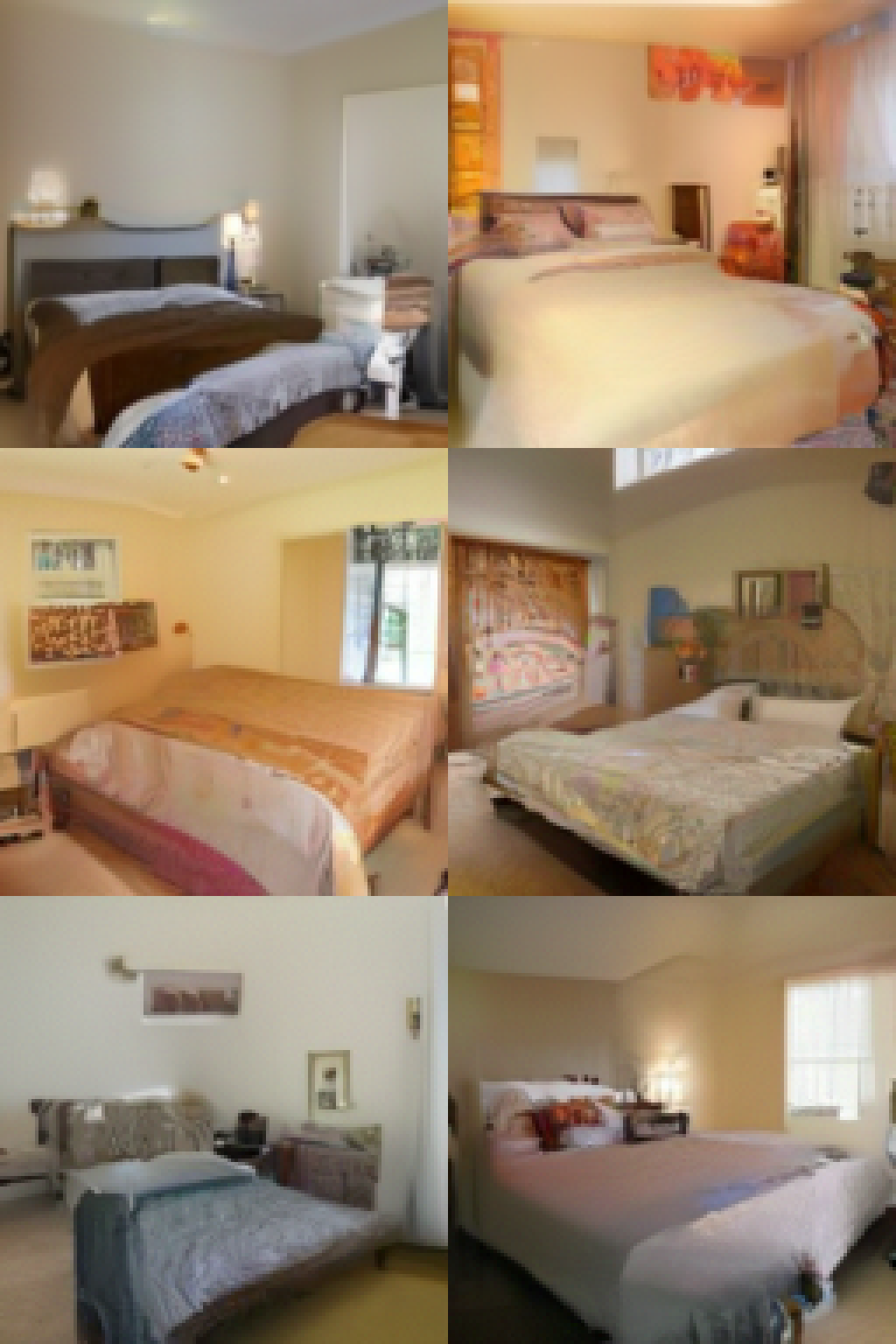}
\includegraphics[width=.325\textwidth]{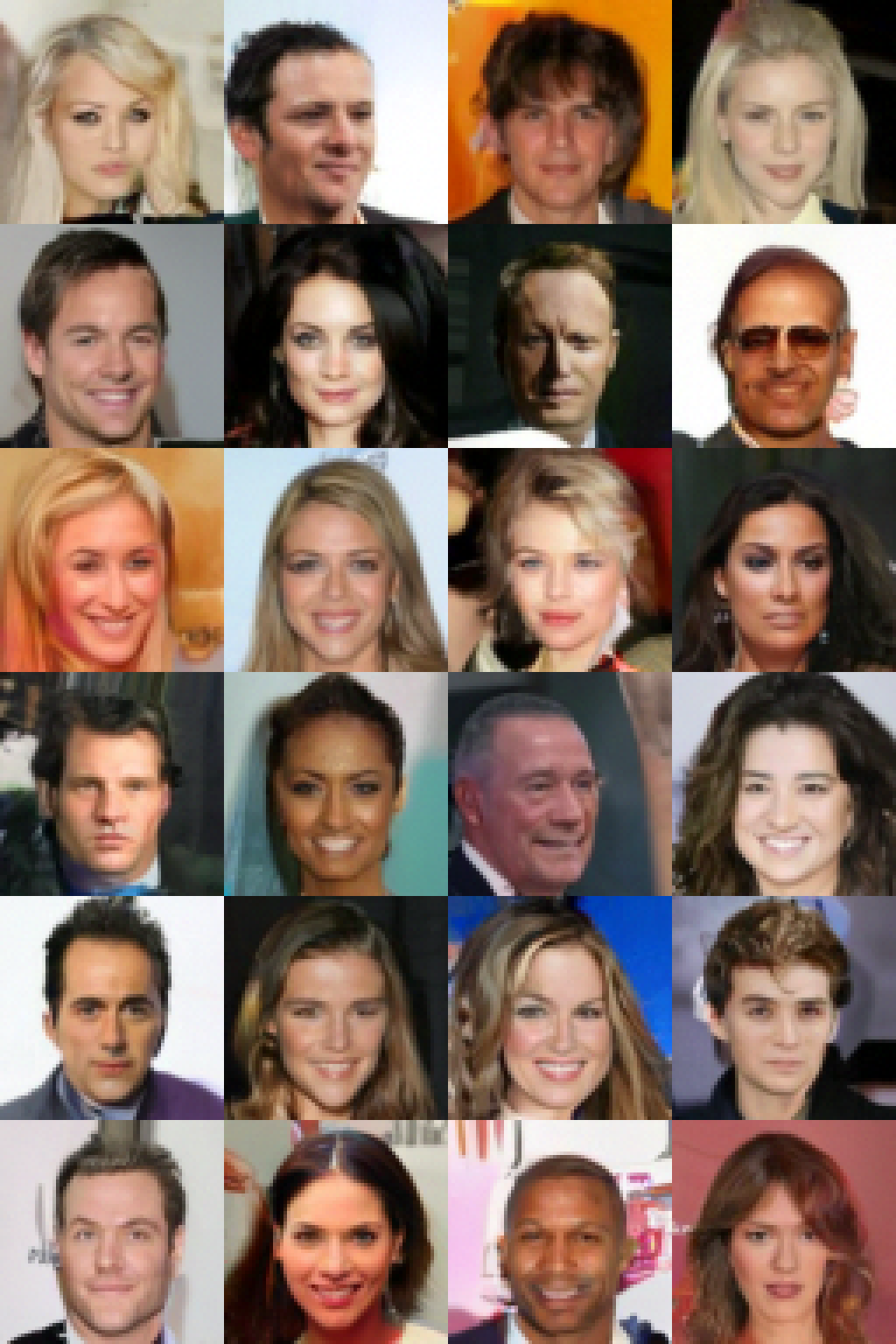} 
\end{center}
\caption{Generated samples on LSUN $128^2$ church\_outdoor ({\em left}), LSUN $128^2$ bedroom ({\em center}) and CelebA $64^2$ ({\em right}).} 
\label{fig:img1}
\end{figure}

\section{Background}
Let $\rvx \sim \pdata(\rvx)$ denote a training example, and $p_\theta(\rvx)$ denote a model's probability density function that aims to approximates $\pdata(\rvx)$. An energy-based model (EBM) is defined as:
\begin{eqnarray} 
     p_\theta(\rvx) = \frac{1}{Z_\theta}\exp(f_\theta(\rvx)), 
     \label{eqn:ebm}
\end{eqnarray}
where $Z_\theta = \int \exp(f_\theta(\rvx)) d\rvx$ is the partition function, which is analytically intractable for high-dimensional $\rvx$. For images, we parameterize $f_\theta(\rx)$ with a convolutional neural network with a scalar output. 

The energy-based model in \eqref{eqn:ebm} can, in principle, be learned through MLE. Specifically, suppose we observe samples $\rvx_i \sim \pdata(\rvx)$ for $i = 1, 2, ..., n$. The log-likelihood function is 
\begin{eqnarray}
	\mathcal{L}(\theta) = \frac{1}{n} \sum_{i=1}^n \log p_\theta(\rvx_i) \doteq \E_{\rvx \sim \pdata} [\log p_\theta(\rvx)]. \label{eqn:mle}
\end{eqnarray}  

In MLE, we seek to maximize the log-likelihood function, where the gradient approximately follows~\citep{xie2016theory}
\begin{eqnarray}
	- \frac{\partial}{\partial \theta} \KL( \pdata \Vert p_\theta) = \E_{\rvx \sim \pdata}\left[\frac{\partial}{\partial \theta} f_\theta(\rvx) \right] - \E_{\rvx \sim p_\theta}\left[\frac{\partial}{\partial \theta} f_\theta(\rvx) \right]. \label{eqn: mle_learning}
\end{eqnarray}
The expectations can be approximated by averaging over the observed samples and the synthesized samples drawn from the model distribution $p_\theta(\rvx)$ respectively. Generating synthesized samples from $p_\theta(\rvx)$ can be done with Markov Chain Monte Carlo (MCMC) such as Langevin dynamics (or Hamiltonian Monte Carlo ~\citep{girolami2011riemann}), which iterates
\begin{eqnarray}
	\rvx^{\tau + 1} = \rvx^\tau + \frac{\delta^2}{2} \nabla_\rvx f_\theta(\rvx^\tau) + \delta \beps^\tau,
\end{eqnarray}
\begin{wrapfigure}{r}{0.39\linewidth}
\centering
\includegraphics[width=.38\textwidth]{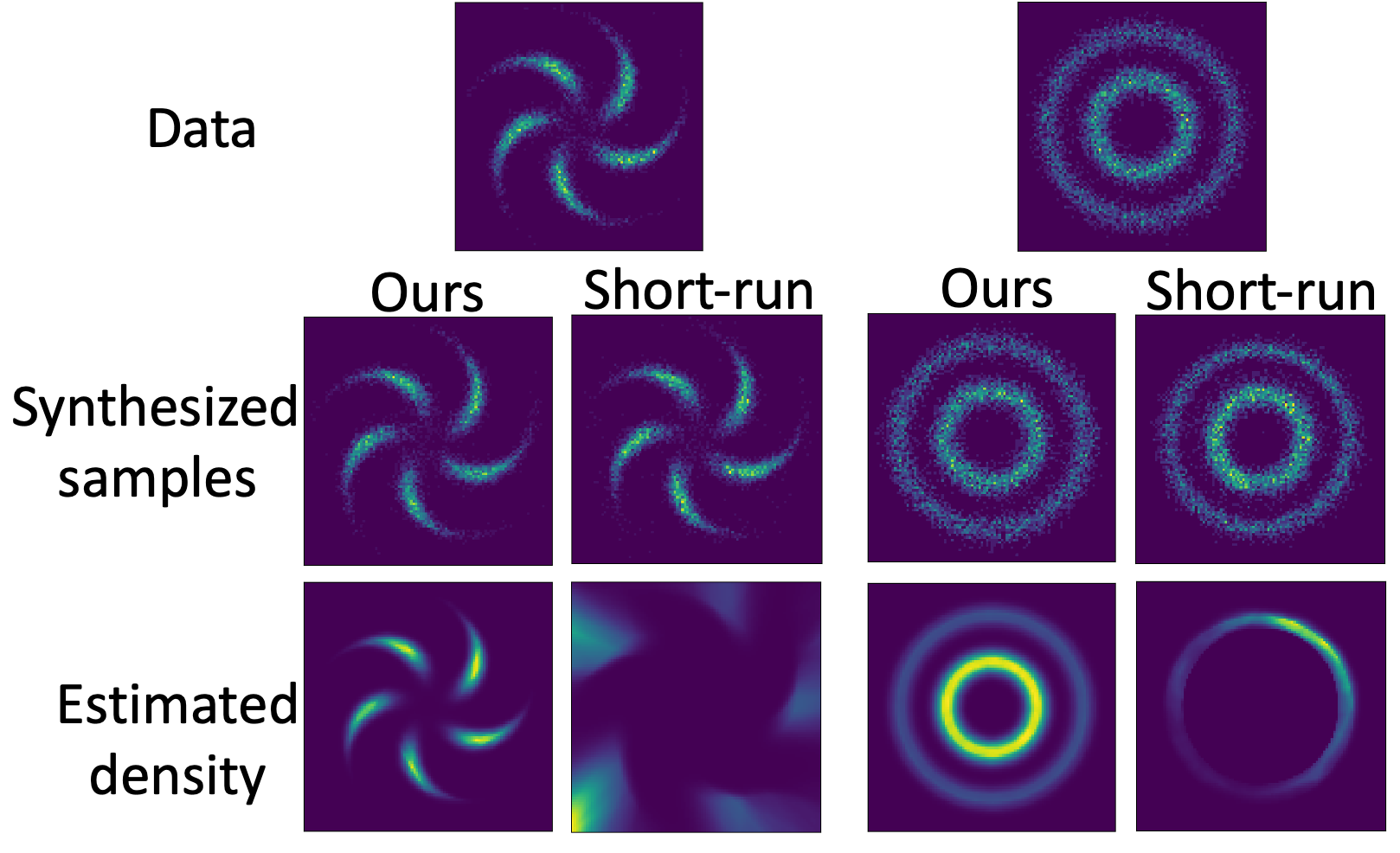}
\caption{Comparison of learning EBMs by diffusion recovery likelihood (Ours) versus marginal likelihood (Short-run).}
\vspace{-.7cm}
\label{fig:2d_comp}
\end{wrapfigure}
where $\tau$ indexes the time, $\delta$ is the step size, and $\beps^\tau \sim \N(0, \mI)$. The difficulty lies in the fact that for high-dimensional and multi-modal distributions, MCMC sampling can take a long time to converge, and the sampling chains may have difficulty traversing modes. As demonstrated in \Figref{fig:2d_comp}, training EBMs with synthesized samples from non-convergent MCMC results in malformed energy landscapes \citep{nijkamp2019learning}, even if the samples from the model look reasonable.  
\section{Recovery Likelihood}
\begin{figure}[t]
\begin{center}
\includegraphics[width=.8\textwidth]{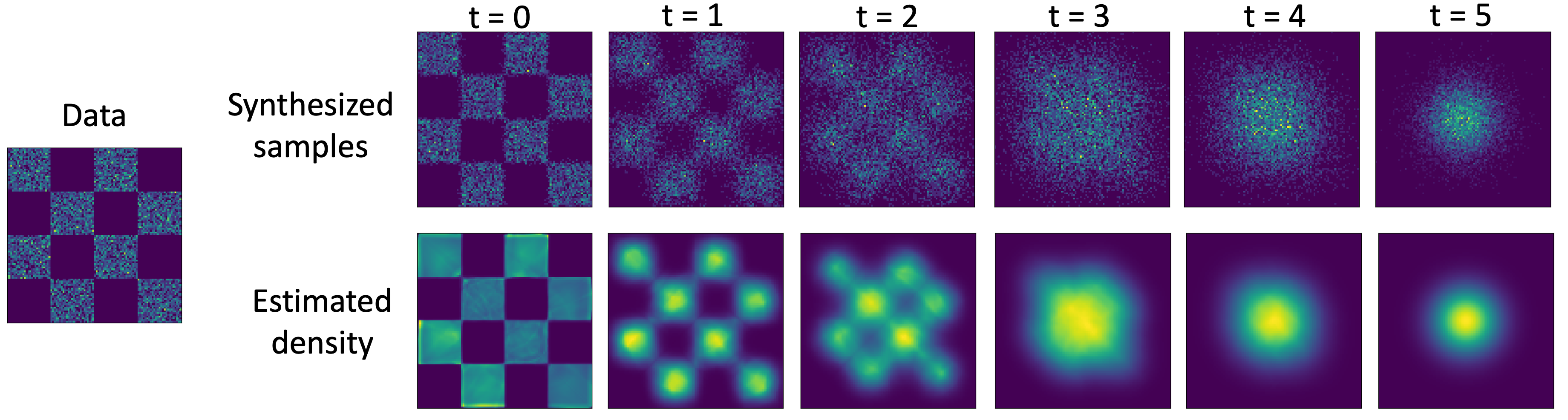}
\end{center}
\caption{Illustration of diffusion recovery likelihood on 2D checkerboard example. {\em Top}: progressively generated samples. {\em Bottom}: estimated marginal densities. }
\label{fig:2d}
\end{figure}
\subsection{From Marginal to Conditional}
Given the difficulty of sampling from the marginal density $p_\theta(\rvx)$, following  \citet{bengio2013generalized}, we use the recovery likelihood defined by the density of the observed sample conditional on a noisy sample perturbed by isotropic Gaussian noise. Specifically, let $\tilde{\rvx} = \rvx + \sigma \beps$ be the noisy observation of $\rvx$, where $\beps \sim \N(0, \mI)$. Suppose $p_\theta(\rvx)$ is defined by the EBM in  \eqref{eqn:ebm}, then the conditional EBM can be derived as 
\begin{eqnarray} 
   p_\theta(\rvx|\trvx) = \frac{1}{\tilde{Z}_\theta(\trvx)} \exp\left(f_\theta(\rvx) - \frac{1}{2\sigma^2} \|\trvx-\rvx\|^2\right),  \label{eq:r}
\end{eqnarray}
where $\tilde{Z}_\theta(\trvx) = \int \exp\left(f_\theta(\rvx) - \frac{1}{2\sigma^2} \|\trvx-\rvx\|^2\right) d\rvx$ is the partition function of this conditional EBM. See Appendix \ref{app:deri} for the derivation. 
Compared to $p_\theta(\rvx)$ (\eqref{eqn:ebm}), the extra quadratic term $\frac{1}{2\sigma^2} \|\trvx-\rvx\|^2$ in $p_\theta(\rvx|\trvx)$ constrains the energy landscape to be localized around $\trvx$, making the latter less multi-modal and easier to sample from. As we will show later, when $\sigma$ is small, $p_\theta(\rvx|\trvx)$ is approximately a single mode Gaussian distribution, which greatly reduces the burden of MCMC. 

A more general formulation is $\trvx = a \rvx + \sigma \beps$, where $a$ is a positive constant. In that case, we can let $\rvy = a \rvx$, and treat $\rvy$ as the observed sample. Assume $p_\theta(\rvy) = \frac{1}{Z_\theta}\exp(f_\theta(\rvy))$, then by \emph{change of variable}, the density function of $\rvx$ can be derived as $g_\theta(\rvx) = a p_\theta(a \rvx)$.

\subsection{Maximizing recovery likelihood} \label{sect: learning}
With the conditional EBM, assume we have observed samples $\rvx_i \sim \pdata(\rvx)$ and the corresponding perturbed samples $\trvx_i = \rvx_i + \sigma \beps_i$ for $i = 1,...,n$. We define the \emph{recovery log-likelihood function} as
\begin{equation}
	\mathcal{J}(\theta) = \frac{1}{n} \sum_{i=1}^n \log p_\theta(\rvx_i|\trvx_i). 
\end{equation}
The term \emph{recovery} indicates that we attempt to recover the clean sample $\rvx_i$ from the noisy sample $\trvx_i$. Thus, instead of maximizing $\mathcal{L}(\theta)$ in \eqref{eqn:mle}, we can maximize $\mathcal{J}({\theta})$, whose distributions are easier to sample from. Specifically, we generate synthesized samples by $K$ steps of Langevin dynamics that iterates
\begin{eqnarray}
	\rvx^{\tau + 1} = \rvx^\tau + \frac{\delta^2}{2} (\nabla_{\rvx}f_\theta(\rvx^\tau) + \frac{1}{\sigma^2} (\trvx-\rvx^\tau)) + \delta \beps^\tau.
\end{eqnarray}
The model is then updated following the same learning gradients as MLE (\eqref{eqn: mle_learning}), because the quadratic term $- \frac{1}{2\sigma^2} \|\trvx-\rvx\|^2$ is not related to $\theta$. Following the classical analysis of MLE, we can show that the point estimate given by maximizing recovery likelihood is an unbiased estimator of the true parameters, which means that given enough data, a rich enough model and exact synthesis, maximizing the recovery likelihood learns $\theta$ such that $\pdata(\rvx) = p_\theta(\rvx)$. See Appendix \ref{app: theo} for a theoretical explanation.

\subsection{Normal Approximation to Conditional}
When the variance of perturbed noise $\sigma^2$ is small, $p_\theta(\rvx|\trvx)$ can be approximated by a normal distribution via a first order Taylor expansion at $\trvx$. Specifically, the negative conditional energy is 
\begin{align} 
 - \mathcal{E}_\theta(\rvx|\trvx) &=   f_\theta(\rvx) - \frac{1}{2\sigma^2} \|\trvx-\rvx\|^2 \\
        &\doteq f_\theta(\trvx) + \langle \nabla_\rvx f_\theta(\trvx), \rvx-\trvx\rangle  - \frac{1}{2\sigma^2} \|\trvx-\rvx\|^2\\  
        &= - \frac{1}{2\sigma^2} \left[ \|\rvx - (\trvx + \sigma^2 \nabla_\rvx f_\theta(\trvx))\|^2\right] + c, \label{eq:normal1}
\end{align}
where ${c}$ include terms irrelevant of $\rvx$ (see Appendix \ref{app:normal} for a detailed derivation). In the above approximation, we do not perform second order Taylor expansion because $\sigma^2$ is small, and $\|\trvx-\rvx\|^2/2\sigma^2$ will dominate all the second order terms from Taylor expansion. 
Thus we can approximate $p_\theta(\rvx|\trvx)$ by a Gaussian approximation $\widetilde{p}_\theta(\rvx|\trvx)$:
\begin{eqnarray} 
    \widetilde{p}_\theta(\rvx|\trvx) = \N\left(\rvx ; \trvx + \sigma^2 \nabla_\rvx f_\theta(\trvx), \sigma^2\right). \label{eqn:normal}
\end{eqnarray}
We can sample from this distribution using: 
\begin{eqnarray} 
     \rvx_{\rm gen} = \trvx + \sigma^2 \nabla_\rvx f_\theta(\trvx) + \sigma \beps,\label{eq:normal_approx}
\end{eqnarray} 
where $\beps \sim \N(0, I)$. This resembles a single step of Langevin dynamics, except that $\sigma \beps$ is replaced by $\sqrt{2} \sigma \beps$ in Langevin dynamics. This normal approximation has two traits: (1) it verifies the fact that the conditional density $p_\theta(\rvx | \trvx)$ can be generally easier to sample from when $\sigma$ is small; (2) it provides hints of choosing the step size of Langevin dynamics, as discussed in \secref{sect:seq}.

\subsection{Connection to variational inference and score matching} \label{sect:connection}
The normal approximation to the conditional distribution leads to a natural connection to diffusion probabilistic models~\citep{sohl2015deep,ho2020denoising} and denoising score matching~\citep{vincent2011connection,song2019generative,song2020improved,saremi2018deep,saremi2019neural}. Specifically, in diffusion probabilistic models, instead of modeling $p_\theta(x)$ as an energy-based model, it recruits variational inference and directly models the conditional density as
\begin{eqnarray}
p_\theta(\rvx|\trvx) \sim  \mathcal{N}\left(\trvx + \sigma^2 s_\theta(\trvx), \sigma^2\right) ,
\end{eqnarray}
which is in agreement with the normal approximation (\eqref{eqn:normal}), with $s_\theta(\rvx) = \nabla_\rvx f_\theta(\rvx)$. On the other hand, the training objective of denoising score matching is to minimize
\begin{eqnarray}
	\frac{1}{2\sigma^2} \E_{p(\rvx, \trvx)}[\|\rvx - ( \trvx  + \sigma^2 s_\theta(\trvx))\|^2],
\end{eqnarray}
where $s_\theta(\rvx)$ is the score of the density of $\trvx$. This objective is in agreement with the objective of maximizing log-likelihood of the normal approximation (\eqref{eq:normal1}), except that for normal approximation, $\nabla_\rvx f_\theta(\cdot)$ is the score of density of $\rvx$, instead of $\trvx$. However, the difference between the scores of density of $\rvx$ and $\trvx$ is of $O(\sigma^2)$, which is negligible when $\sigma$ is sufficiently small (see Appendix \ref{app:diff} for details). We can further show that the learning gradient of maximizing log-likelihood of the normal approximation is approximately the same as the learning gradient of maximizing the original recovery log-likelihood with one step of Langevin dynamics (see Appendix \ref{app:grad}). It indicates that the training process of maximizing recovery likelihood agrees with the one of diffusion probabilistic models and denoising score matching when $\sigma$ is small. 

As the normal approximation is accurate only when $\sigma$ is small, it requires many time steps in the diffusion process for this approximation to work well, which is also reported in~\citet{ho2020denoising} and ~\citet{song2020improved}. In contrast, the diffusion recovery likelihood framework can be more flexible in choosing the number of time steps and the magnitude of $\sigma$.

\subsection{Diffusion recovery likelihood} \label{sect:seq}
As we discuss, sampling from $p_\theta(\rvx | \trvx)$ becomes simple only when $\sigma$ is small. In the extreme case when $\sigma \rightarrow \infty$, $p_\theta(\rvx | \trvx)$ converges to the marginal distribution $p_\theta(\rvx)$, which is again highly multi-modal and difficult to sample from. To keep $\sigma$ small and meanwhile equip the model with the ability to generate new samples initialized from white noise, inspired by~\citet{sohl2015deep} and~\citet{ho2020denoising}, we propose to learn a sequence of recovery likelihoods, on gradually perturbed observed data based on a diffusion process. Specifically, assume a sequence of perturbed observations $\rvx_0, \rvx_1, ..., \rvx_T$ such that 
\begin{eqnarray}
	\rvx_0 \sim \pdata(\rvx);\;\; \rvx_{t+1} = \sqrt{1 - \sigma_{t+1}^2} \rvx_{t} + \sigma_{t+1} \beps_{t+1},\;\;  t = 0, 1,... T-1.
\end{eqnarray}
The scaling factor $\sqrt{1 - \sigma_{t+1}^2}$ ensures that the sequence is a spherical interpolation between the observed sample and Gaussian white noise. Let $\rvy_{t} = \sqrt{1 - \sigma_{t+1}^2} \rvx_{t}$, and we assume a sequence of conditional EBMs
\begin{eqnarray} 
   p_\theta(\rvy_{t}|\rvx_{t+1}) = \frac{1}{\tilde{Z}_{\theta, t}(\rvx_{t+1})} \exp\left(f_\theta(\rvy_{t}, t) - \frac{1}{2\sigma_{t+1}^2} \|\rvx_{t+1}-\rvy_t\|^2\right), \;\; t = 0, 1, ..., T-1,
\end{eqnarray}
where $f_\theta(\rvy_t, t)$ is defined by a neural network conditioned on $t$. 

We follow the learning algorithm in \secref{sect: learning}. A question is how to determine the step size schedule $\delta_t$ of Langevin dynamics. Inspired by the sampling procedure of the normal approximation (\eqref{eq:normal_approx}), we set the step size $\delta_t = b \sigma_t$, where $b < 1$ is a tuned hyperparameter. This schedule turns out to work well in practice. Thus the $K$ steps of Langevin dynamics iterates
\begin{eqnarray}
	\rvy^{\tau+1}_t = \rvy^{\tau}_t + \frac{b^2\sigma_t^2}{2} (\nabla_\rvy f_\theta(\rvy^\tau_t, t) + \frac{1}{\sigma_t^2} (\rvx_{t+1}-\rvy_t^\tau)) + b \sigma_t \beps^\tau. \label{eqn:sampling}
\end{eqnarray}
\Algref{algo:1} summarizes the training procedure. After training, we initialize the MCMC sampling from Gaussian white noise, and the synthesized sample at each time step serves to initialize the MCMC that samples from the model of the previous time step. See \algref{algo:2}. To show the efficacy of our method, Figures \ref{fig:2d} and \ref{fig:2d_comp} display several 2D toy examples learned by diffusion recovery likelihood. 
\vspace{-.5cm}
\begin{center}
	\begin{minipage}[c]{.45\textwidth}
	\begin{algorithm}[H]
	\caption{Training}
	\label{algo:1}	
\begin{algorithmic}
	\REPEAT
	\STATE Sample $t \sim {\rm Unif}(\{0, ..., T-1\})$.
	\STATE Sample pairs $(\rvy_{t}, \rvx_{t+1})$.
	\STATE Set synthesized sample $\rvy_t^- = \rvx_{t+1}$.
	\FOR{$\tau \leftarrow 1$ to $K$}
	\STATE Update $\rvy_t^-$ according to \eqref{eqn:sampling}.
	\ENDFOR
	\STATE Update $\theta$ following the gradients \\
	$\frac{\partial}{\partial \theta} f_\theta(\rvy_t, t)- \frac{\partial}{\partial \theta} f_\theta(\rvy_t^-, t)$.
	\UNTIL converged.
\end{algorithmic}
\end{algorithm}
\end{minipage}
\hspace{3mm}
\begin{minipage}[c]{.45\textwidth}
	\begin{algorithm}[H]
	\caption{Progressive sampling}
	\label{algo:2}	
\begin{algorithmic}
	\STATE Sample $\rvx_T \sim \N(0, \mI)$.
	\FOR{$t \gets T-1$ to $0$}
	\STATE $\rvy_t = \rvx_{t+1}$.
	\FOR{$\tau \leftarrow 1$ to $K$}
	\STATE Update $\rvy_t$ according to \eqref{eqn:sampling}.
	\ENDFOR
	\STATE $\rvx_t = \rvy_t / \sqrt{1 - \sigma_{t+1}^2}$.
	\ENDFOR
	\RETURN $\rvx_0$.
\end{algorithmic}
\end{algorithm}
\end{minipage}
\end{center}

\section{Experiments}
To show that diffusion recovery likelihood is flexible for diffusion process of various magnitudes of noise, we test the method under two settings: (1) $T=6$, with $K=30$ steps of Langevin dynamic per time step; (2) $T = 1000$, with sampling from normal approximation. (2) resembles the noise schedule of~\citet{ho2020denoising} and the magnitude of noise added at each time step is much smaller compared to (1). For both settings, we set $\sigma_t^2$ to increase linearly. The network structure of $f_\theta(x, t)$ is based on Wide ResNet~\citep{zagoruyko2016wide} and we remove weight normalization. $t$ is encoded by Transformer sinusoidal position embedding as in~\citep{ho2020denoising}. For (1), we find that including another scaling factor $c_t$ to the step size $\delta_t$ helps. Architecture and training details are in Appendix \ref{app:exp}. Henceforth we simply refer the two settings as {\em T6} and {\em T1k}. 

\subsection{Image generation}
\begin{figure}[ht]
\begin{center}
\includegraphics[width=.325\textwidth]{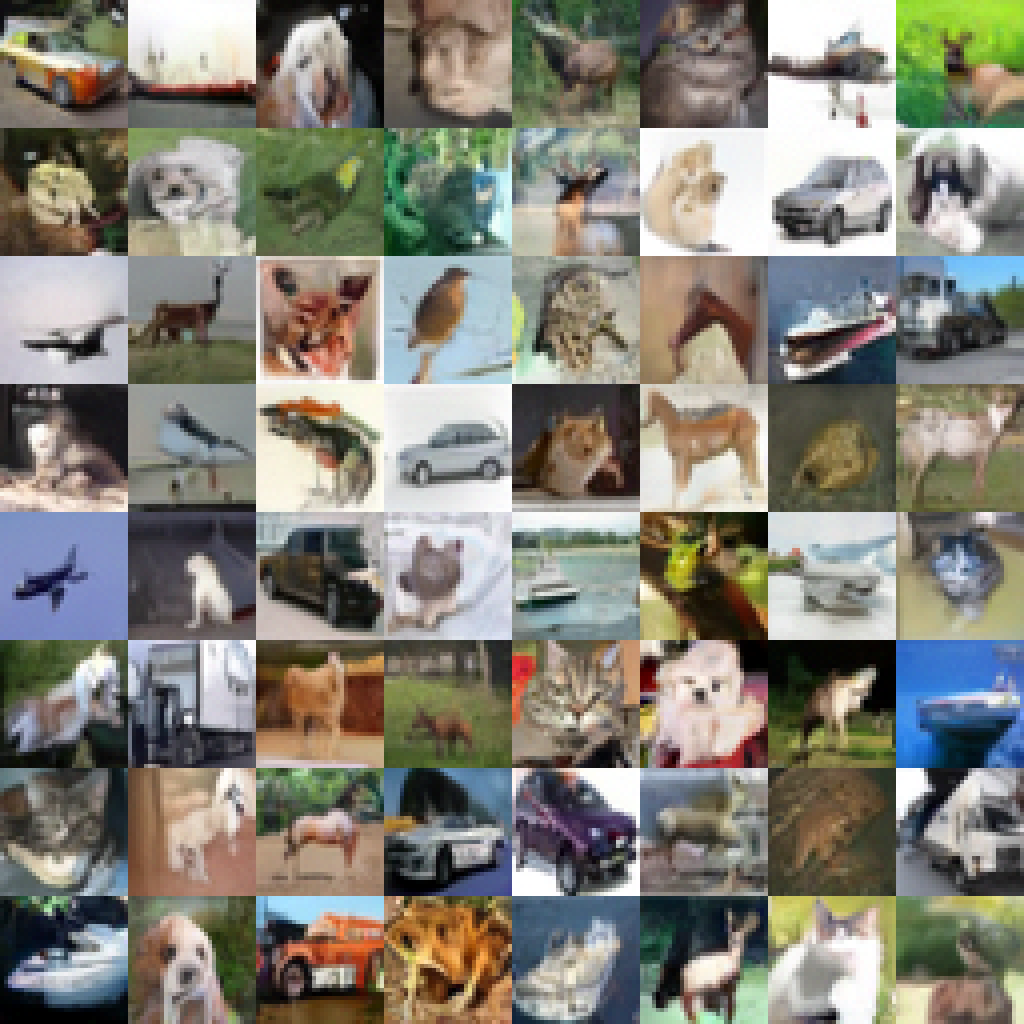}
\includegraphics[width=.325\textwidth]{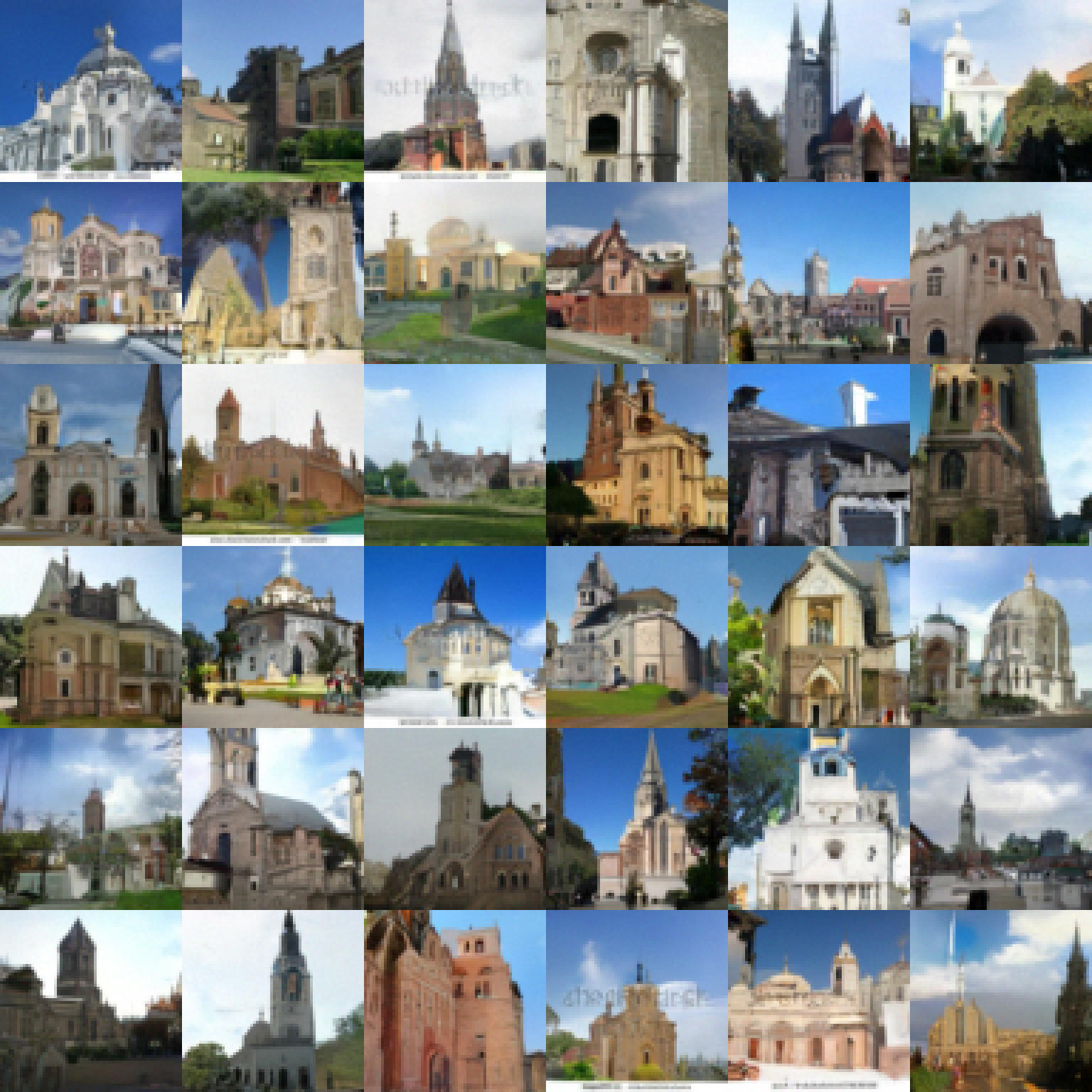}
\includegraphics[width=.325\textwidth]{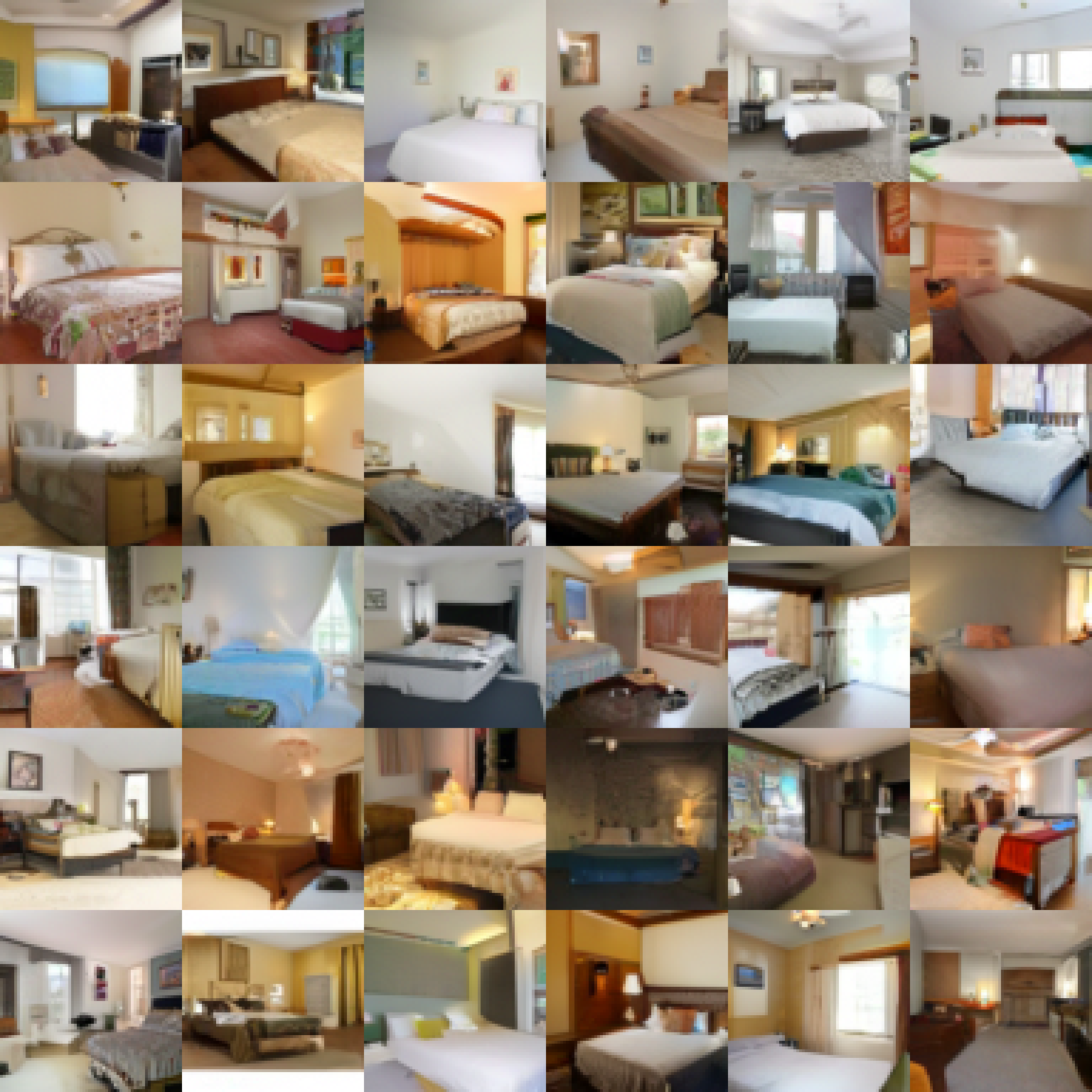}
\end{center}
\caption{Generated samples on unconditional CIFAR-10 ({\em left}) and LSUN $64^2$ church\_outdoor ({\em center}) and LSUN $64^2$ bedroom ({\em right}).}
\label{fig:img2}
\end{figure}

Figures \ref{fig:img1} and \ref{fig:img2} display uncurated samples generated from learned models on CIFAR-10, CelebA $64 \times 64$, LSUN $64 \times 64$ and $128 \times 128$ datasets under {\em T6} setting. The samples are of high fidelity and comparable to GAN-based methods. Appendix \ref{app:samples} provides more generated samples. Tables \ref{tabl: fid} and \ref{tabl: fid-celeba} summarize the quantitative evaluations on CIFAR-10 and CelebA datasets, in terms of Frechet Inception Distance (FID)~\citep{heusel2017gans} and inception scores~\citep{salimans2016improved}. On CIFAR-10, our model achieves FID 9.58 and inception score 8.30, which outperforms existing methods of learning explicit energy-based models to a large extent, and is superior to a majority of GAN-based methods. On CelebA, our model obtains results comparable with the state-of-the-art GAN-based methods, and outperforms score-based methods~\citep{song2019generative,song2020improved}.  Note that the score-based methods \citep{song2019generative,song2020improved} and diffusion probabilistic models~\citep{ho2020denoising} directly parametrize and learn the score of data distribution, whereas our goal is to learn explicit energy-based models. 

\begin{center}
\begin{minipage}[c]{.55\textwidth}
\begin{table}[H]
\centering
 \caption{FID and inception scores on CIFAR-10. } 
 \footnotesize
 \setlength\tabcolsep{3.0pt}
 \begin{tabular}{lcc}
    \toprule
    Model & FID$\downarrow$ & Inception$\uparrow$  \\
        \midrule
      {\bf GAN-based} & & \\
      \midrule
      WGAN-GP~\citep{gulrajani2017improved} & $36.4$ & 7.86 $\pm$ .07 \\
      SNGAN~\citep{miyato2018spectral} & 21.7 & 8.22 $\pm$ .05 \\
      SNGAN-DDLS~\citep{che2020your} & 15.42 & 9.09 $\pm$ .10 \\
      StyleGAN2-ADA~\citep{karras2020training} & 3.26 & {\bf 9.74} $\pm$ .05 \\
      \midrule
      {\bf Score-based} & &\\
      \midrule
      NCSN~\citep{song2019generative} & 25.32 & 8.87 $\pm$ .12 \\
      NCSN-v2~\citep{song2020improved} & 10.87 & 8.40 $\pm$ .07 \\
      DDPM~\citep{ho2020denoising} & {\bf 3.17} & 9.46 $\pm$ .11\\
    \midrule 
      {\bf Explicit EBM-conditional} & & \\
      \midrule
      CoopNets~\citep{xie2019cooperative} & - & 7.30 \\
      EBM-IG~\citep{du2019implicit} & 37.9 & 8.30 \\
      JEM~\citep{grathwohl2019your} & 38.4 & 8.76 \\
      \midrule
      {\bf Explicit EBM} & & \\
      \midrule
           Muli-grid~\citep{gao2018learning} & 40.01 & 6.56\\
      CoopNets~\citep{xie2016cooperative} & 33.61 & 6.55 \\
      EBM-SR~\citep{nijkamp2019learning} & - & 6.21 \\
      EBM-IG~\citep{du2019implicit} & 38.2 & 6.78 \\
      {\bf Ours} ({\em T6}) & {\bf 9.58} & {\bf 8.30} $\pm$ .11 \\ 
        \bottomrule 
    \end{tabular}
    \label{tabl: fid}
\end{table}
\end{minipage}
\begin{minipage}[c]{.44\textwidth}
\begin{table}[H]
\centering
	\caption{Ablation of training objectives, time steps $T$ and sampling steps $K$ on CIFAR-10. $K = 0$ indicates that we sample from the normal approximation.}  
	\footnotesize
	 \setlength\tabcolsep{4 pt}
 \begin{tabular}{lcc} 
    \toprule
    Setting / Objective & FID$\downarrow$ & Inception$\uparrow$ \\
    \midrule
    T = 1, K = 180 & 32.12 & 6.72 $\pm$ 0.12 \\
    T = 1000, K = 0 & 22.58 & 7.71 $\pm$ 0.08 \\
    T = 1000, K = 0 (DSM) & 21.76 & 7.76 $\pm$ 0.11 \\
    T = 6, K = 10 & - & - \\
    T = 6, K = 30 & 9.58 & 8.30 $\pm$ 0.11 \\
    T = 6, K = 50 & {\bf 9.36} & {\bf 8.46} $\pm$ 0.13 \\
        \bottomrule 
    \end{tabular}
    \label{tabl: ablation}
\end{table}
\vspace{-4mm}
\begin{table}[H]
\centering
	\caption{FID scores on CelebA $64^2$.} 
	\vspace{-0.2cm}
	\footnotesize
	 \setlength\tabcolsep{1.5pt}
	 \begin{adjustbox}{max width=\textwidth}
     \begin{tabular}{lc} 
        \toprule
        Model & FID$\downarrow$ \\
        \midrule
        QA-GAN~\citep{parimala2019quality} & 6.42 \\
        COCO-GAN~\citep{lin2019coco} & {\bf 4.0} \\
        \midrule
        NVAE~\citep{vahdat2020nvae} & 14.74 \\
        \midrule
        NCSN~\citep{song2019generative} & 25.30 \\
        NCSN-v2~\citep{song2020improved} & 10.23 \\
        \midrule
        EBM-SR~\citep{nijkamp2019learning} & 23.02\\
        EBM-Triangle~\citep{han2020joint} & 24.70 \\
        {\bf Ours} ({\em T6}) & 5.98 \\
            \bottomrule 
    \end{tabular}
    \end{adjustbox}
    \label{tabl: fid-celeba}
\end{table}
\end{minipage}
\end{center}
\begin{figure}[ht]
\begin{center}
\includegraphics[width=\textwidth]{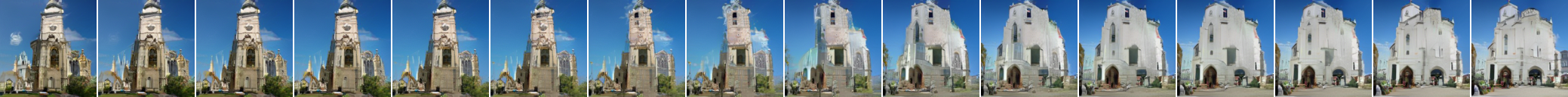} 
\includegraphics[width=\textwidth]{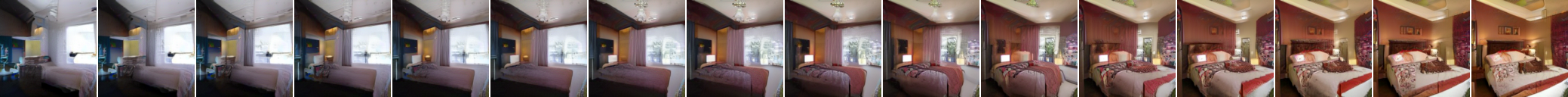}
\includegraphics[width=\textwidth]{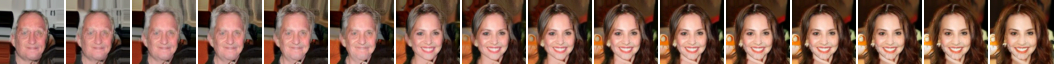}
\end{center}
\caption{Interpolation results between the leftmost and rightmost generated samples. For {\em top} to {\em bottom}: LSUN church\_outdoor $128^2$, LSUN bedroom $128^2$ and CelebA $64^2$. } 
\label{fig: interp}
\end{figure}
\begin{center}
	\begin{minipage}[c]{.46\textwidth}
		\begin{table}[H]
\centering
	\caption{Test bits per dimension on CIFAR-10. $^\dagger$ indicates that we estimate the bit per dimension with the approximated log partition function instead of analytically computing it. See section \ref{sect:long-run}. } 
	\footnotesize
	 \setlength\tabcolsep{2.5pt}
 \begin{tabular}{lc} 
    \toprule
    Model & BPD$\downarrow$ \\
    \midrule
    DDPM~\citep{ho2020denoising} & 3.70 \\
    Glow~\citep{kingma2018glow} & 3.35 \\
    Flow++~\citep{ho2019flow++} & 3.08 \\
    GPixelCNN~\citep{van2016conditional} & 3.03 \\
    Sparse Transformer~\citep{child2019generating} & 2.80 \\
    DistAug~\citep{icml2020_6095} & {\bf 2.56} \\
    {\bf Ours}$^\dagger$ ({\em T1k}) & 3.18 \\
        \bottomrule 
    \end{tabular}
    \label{tabl: bpd}
\end{table}
	\end{minipage}
	\begin{minipage}[c]{.53\textwidth}
\begin{figure}[H]
\begin{center}
\includegraphics[width=.49\textwidth]{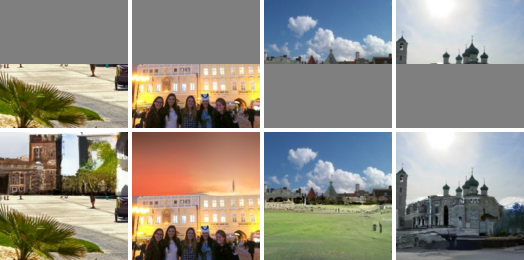} 
\includegraphics[width=.49\textwidth]{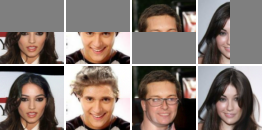}\\
\vspace{1mm}
\includegraphics[width=.49\textwidth]{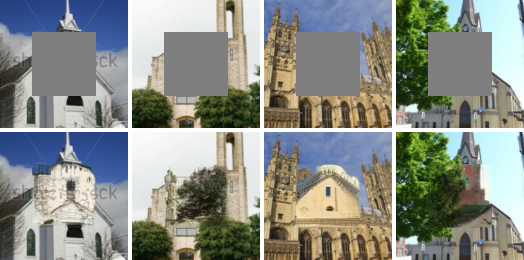} 
\includegraphics[width=.49\textwidth]{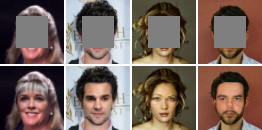}\\
\end{center}
\caption{Image inpainting on LSUN church\_outdoor $128^2$ ({\em left}) and CelebA $64^2$ ({\em right}). With each block, the top row are mask images while the bottom row are inpainted images.} 
\label{fig: inpaint}
\end{figure} 
	\end{minipage}
\end{center}

\paragraph{Interpolation.} As shown in \Figref{fig: interp}, our model is capable of smooth interpolation between two generated samples. Specifically, for two samples $\rvx_0^{(0)}$ and $\rvx_0^{(1)}$, we do a sphere interpolation between the initial white noise images $\rvx_T^{(0)}$ and $\rvx_T^{(1)}$ and the noise terms of Langevin dynamics $\beps_{t, \tau}^{(0)}$ and $\beps_{t, \tau}^{(1)}$ for every sampling step at every time step. More interpolation results can be found in Appendix \ref{app:interp}. 

\paragraph{Image inpainting.} A promising application of energy-based models is to use the learned model as a prior model for image processing, such as image inpainting, denoising and super-resolution \citep{gao2018learning, du2019implicit, song2019generative}. In \Figref{fig: inpaint}, we demonstrate that the learned models by maximizing recovery likelihoods are capable of realistic and semantically meaningful image inpainting. Specifically, given a masked image and the corresponding mask, we first obtain a sequence of perturbed masked images at different noise levels. The inpainting can be easily achieved by running Langevin dynamics progressively on the masked pixels while keeping the observed pixels fixed at decreasingly lower noise levels. Additional image inpainting results can be found in Appendix \ref{app:inpaint}.

\paragraph{Ablation study.}
Table \ref{tabl: ablation} summarizes the results of ablation study on CIFAR-10. We investigate the effect of changing the numbers of time steps $T$ and sampling steps $K$. First, to show that it is beneficial to learn by diffusion recovery likelihood, we compare against a baseline approach ($T=1, K = 180$) where we use only one time step, so that the recovery likelihood becomes marginal likelihood. The approach is adopted by \citet{nijkamp2019learning} and \citet{du2019implicit}. For fair comparison, we equip the baseline method the same budget of MCMC sampling as our {\em T6} setting (i.e., 180 sampling steps). Our method outperforms this baseline method by a large margin. Also the models are trained more efficiently as the number of sampling steps per iteration is reduced and amortized by time steps.

Next, we report the sample quality of setting {\em T1k}. We test two training objectives for this setting: (1) maximizing recovery likelihoods (T = 1000, K = 0) and (2) maximizing the approximated normal distributions (T=1000, K=0 (DSM)). As mentioned in \secref{sect:connection}, (2) is equivalent to the training objectives of denoising score matching~\citep{song2019generative,song2020improved} and diffusion probabilistic model~\citep{ho2020denoising}, except that the score functions are taken as the gradients of explicit energy functions. In practice, for a direct comparison, (2) follows the same implementation as in~\cite{ho2020denoising}, except that the score function is parametrized as the gradients of the explicit energy function used in our method. (1) and (2) achieve similar sample quality in terms of quantitative metrics, where (2) results in a slightly better FID score yet a slightly worse inception score. This verifies the fact that the training objectives of (1) and (2) are consistent. Both (1) and (2) performs worse than setting T6. A possible explanation is that the sampling error may accumulate with many time steps, so that a more flexible schedule of time steps accompanied with certain amount of sampling  steps is preferred. 

Last, we examine the influence of varying the number of sampling steps while fixing the number of time steps. The training becomes unstable when the number of sampling steps are not enough ($T=6, K=10$), and more sampling steps lead to better sample quality. However, since $K = 50$ does not gain significant improvement versus $K = 30$, yet of much higher computational cost, we keep $K = 30$ for image generation on all datasets. See Appendix \ref{app:fid} for a plot of FID scores over iterations. 

\subsection{Long-run chain analysis} \label{sect:long-run}
Besides achieving high quality generation, a perhaps equally important aspect of learning EBMs is to obtain a faithful energy potential. A principle way to check the validity of the learned potential is to perform long-run sampling chains and see if the samples still remain realistic. However, as pointed out in \citet{nijkamp2019anatomy}, almost all existing methods of learning EBMs fail in getting realistic long-run chain samples.  In this subsection, we demonstrate that by composing a thousand diffusion time steps ({\em T1k} setting), we can form steady long-run MCMC chains for the conditional distributions.  
 
\begin{figure}[ht]
\begin{center}
\includegraphics[width=.325\textwidth]{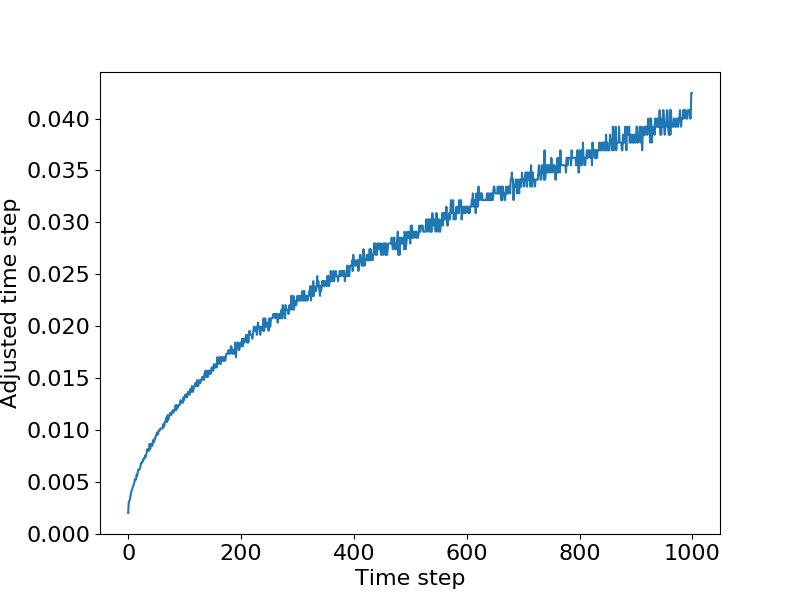}
\includegraphics[width=.325\textwidth]{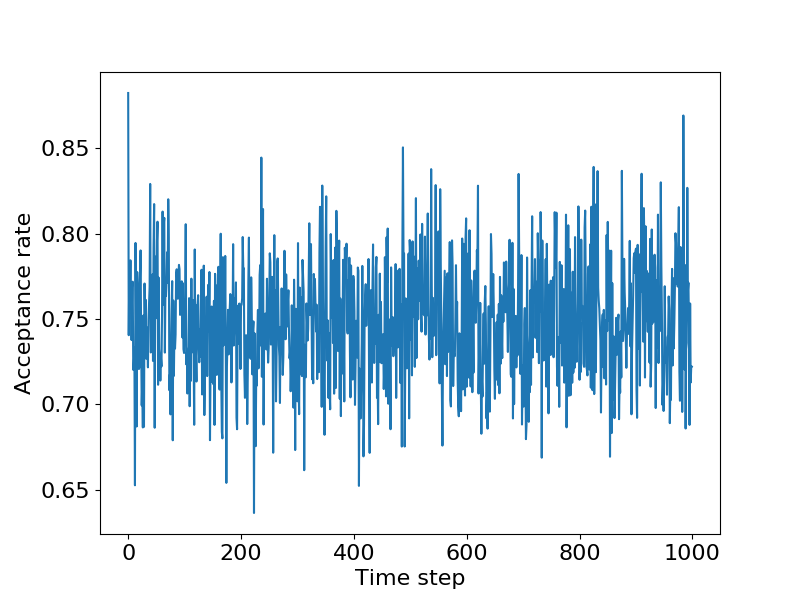}
\includegraphics[width=.325\textwidth]{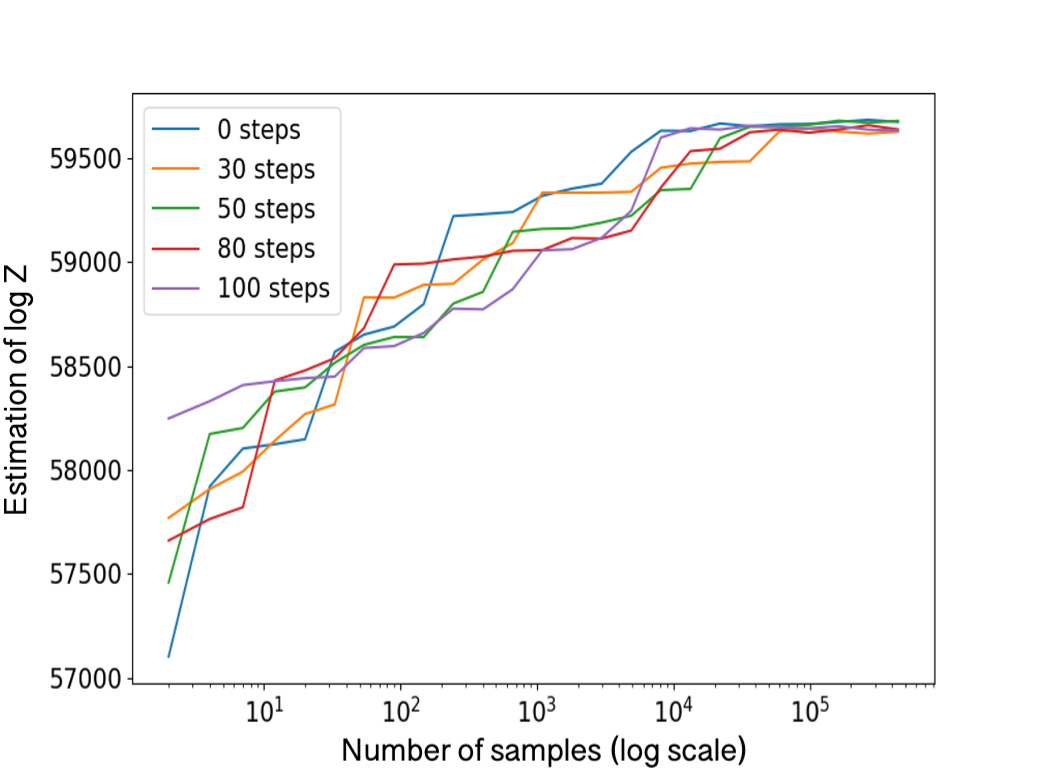}
\end{center}
\caption{{\em Left}: Adjusted step size of HMC over time step. {\em Center}: Acceptance rate over time step. {\em Right}: Estimated log partition function over number of samples with different number of sampling steps per time step. The x axis is plotted in log scale.}
\vspace{-0.3cm}
\label{fig: long-run}
\end{figure}
\begin{wrapfigure}{r}{0.38\linewidth}
\centering
\includegraphics[width=.12\textwidth]{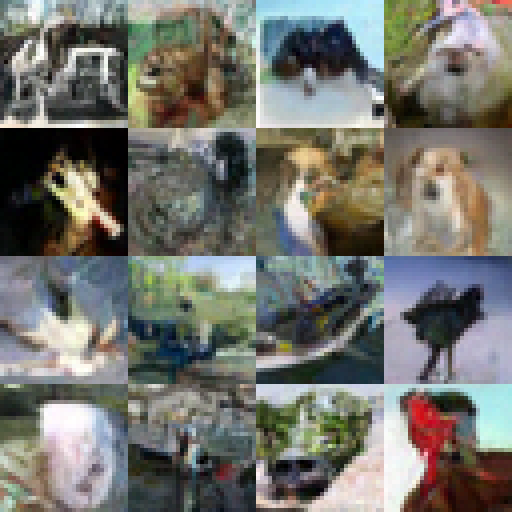}
\includegraphics[width=.12\textwidth]{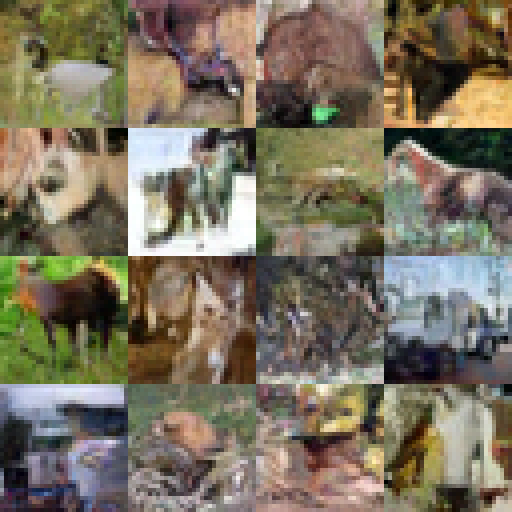}
\includegraphics[width=.12\textwidth]{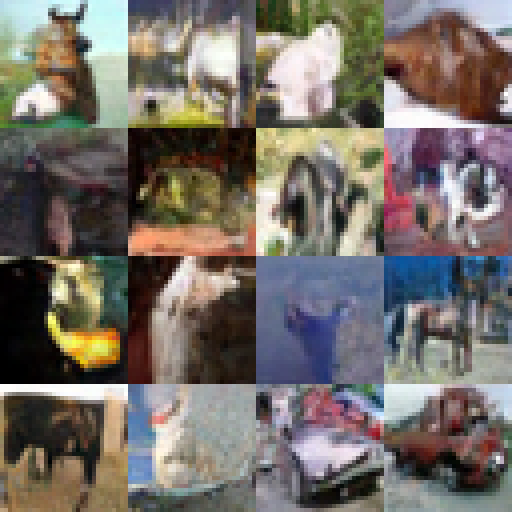}
\caption{Long-run chain samples from model-{\em T1k} with different total amount of HMC steps. From {\em left} to {\em right}: $1k$ steps, $10k$ steps and $100k$ steps.}
\vspace{-.3cm}
\label{fig: long-run-samples}
\end{wrapfigure}

First we prepare a faithful sampler for conducting long-run sampling. Specifically, after training the model under $T1k$ setting by maximizing diffusion recovery likelihood, for each time step, we first sample from the normal approximation and count it as one sampling step, and then use Hamiltonian Monte Carlo (HMC)~\citep{neal2011mcmc} with 2 leapfrog steps to perform the consecutive sampling steps. To obtain a reasonable schedule of sampling step size, for each time step we adaptively adjust the step size of HMC to make the average acceptance rate range in $[0.6, 0.9]$, which is computed over $1000$ chains for $100$ steps. \Figref{fig: long-run} displays the adjusted step size ({\em left}) and acceptance rate ({\em center}) over time step. The adjusted step size increases logarithmically. With this step size schedule, we generate long-run chains from the learned sequence of conditional distributions. As shown in \Figref{fig: long-run-samples}, images remain realistic for even $100k$ sampling steps in total (i.e., $100$ sampling steps per time step), resulting in FID 24.89. This score is close to the one computed on samples generated by $1k$ steps (i.e., sampled from normal approximation), which is 25.12. As a further check, we recruit a No-U-Turn Sampler~\citep{hoffman2014no} with the same step size schedule as HMC to perform long-run sampling, where the samples also remain realistic. See Appendix \ref{app:long-run} for details.

More interestingly, given the faithful long-run MCMC samples from the {\em conditional} distributions, we can estimate the log ratio of the partition functions of the {\em marginal} distributions, and further estimate the partition function of $p_\theta(\rvy_0)$. The strategy is based on annealed importance sampling~\citep{neal2001annealed}. See Appendix \ref{app:ais} for the implementation details. The right subfigure of \Figref{fig: long-run} depicts the estimated log partition function of $p_\theta(\rvy_0)$ over the number of MCMC samples used. To verify the estimation strategy and again check the long-run chain samples, we conduct multiple runs using samples generated with different numbers of HMC steps and display the estimation curves. All the curves saturate to values close to each other at the end, indicating the stability of long-run chain samples and the effectiveness of the estimation strategy. With the estimated partition function, by {\em change of variable}, we can estimate the normalized density of data as $g_\theta(\rvx_0) = \sqrt{1 - \sigma_1^2} p_\theta(\sqrt{1 - \sigma_1^2} \rvx_0)$. We report test bits per dimension on CIFAR-10 in Table \ref{tabl: bpd}. Note that the result should be taken with a grain of salt, because the partition function is estimated by samples and as shown in Appendix \ref{app:ais}, it is a stochastic lower bound of the true value, that will converge to the true value when the number of samples grows large. 

\section{Conclusion} 

We propose to learn EBMs by diffusion recovery likelihood, a variant of MLE applied to diffusion processes. We achieve high quality image synthesis, and with a thousand noise levels, we obtain faithful long-run MCMC samples that indicate the validity of the learned energy potentials. Since this method can learn EBMs efficiently with small budget of MCMC, we are also interested in scaling it up to higher resolution images and investigating this method in other data modalities in the future. 

\section*{Acknowledgement}
The work was done while Ruiqi Gao and Yang Song were interns at Google Brain during the summer of 2020. The work of Ying Nian Wu is supported by NSF DMS-2015577. We thank Alexander A. Alemi, Jonathan Ho, Tim Salimans and Kevin Murphy for their insightful discussions during the course of this project. 

\bibliography{iclr2021_conference}
\bibliographystyle{iclr2021_conference}

\newpage 
\appendix
\section{Extended derivations}
\subsection{Derivation of equation \ref{eq:r}} \label{app:deri}
Let $\tilde{\rvx} = \rvx + \sigma \beps$, where $\beps \sim \N(0, \mI)$. Given the marginal distribution of 
\begin{eqnarray} 
     p_\theta(\rvx) = \frac{1}{Z_\theta}\exp(f_\theta(\rvx)), 
     \label{eqn:ebm-app}
\end{eqnarray}
We can derive the conditional distribution of $\rvx$ given $\trvx$ as
\begin{align}
    p_\theta(\rvx | \trvx) & = p_\theta(\rvx) p(\trvx | \rvx) / p(\trvx) \\
    & = \frac{1}{Z_\theta}\exp(f_\theta(\rvx)) \frac{1}{(2\pi\sigma^2)^{\frac{n}{2}}} \exp(- \frac{1}{2\sigma^2} \|\trvx-\rvx\|^2) / p(\trvx)\\
    & = \frac{1}{\tilde{Z}_\theta(\trvx)} \exp\left(f_\theta(\rvx) - \frac{1}{2\sigma^2} \|\trvx-\rvx\|^2\right),
\end{align}
where we absorb all the terms that are irrelevant of $\rvx$ as $\tilde{Z}_\theta(\trvx)$.

\subsection{Theoretical understanding} \label{app: theo}
In this subsection, we analyze the asymptotic behavior of maximizing the recovery log-likelihood.

For model class $\{p_\theta(\rvx), \forall \theta \}$, suppose there exists $\theta^*$ such that $\pdata = p_{\theta^*}$. According to the classical theory of MLE, let $\hat{\theta}_0$ be the point estimate by MLE. Then we have $\hat{\theta}$ is an unbiased estimator of $\theta^*$ with asymptotic normality: 
\begin{eqnarray} 
   \sqrt{n} (\hat{\theta}_0 - \theta^*) \to \N(0, \mathcal{I}_0(\theta^*)^{-1}), 
\end{eqnarray} 
where $\mathcal{I}_0(\theta) = \E_{\rvx \sim p_\theta} [ - \nabla^2_\theta \log p_\theta(\rvx) ]$  is the Fisher information, and $n$ is the number of observed samples.

Let $\hat{\theta}$ be the point estimate given by maximizing recovery log-likelihood, we can derive a result in parallel to that of MLE:
\begin{eqnarray}
	\sqrt{n} (\hat{\theta} - \theta^*) \to \N(0, \mathcal{I}(\theta^*)^{-1}),
\end{eqnarray}
where $\mathcal{I}(\theta) = \E_{p_{\theta}(\rvx, \trvx)} [ - \nabla^2_\theta \log p_\theta(\rvx|\trvx) ]$. 
The relationship between $I_0(\theta)$ and $I(\theta)$ is that
\begin{eqnarray} 
   \mathcal{I}_0(\theta) = \mathcal{I}(\theta) +  \E_{p_{\theta}(\rvx, \trvx)} [ - \nabla^2_\theta \log p_\theta(\trvx) ].
 \end{eqnarray}
 Thus there is loss of information, but $\hat{\theta}$ is still an unbiased estimator of $\theta^*$ with asymptotic normality.

 \subsection{Detailed derivation of normal approximation} \label{app:normal}
 \begin{align} 
 - \mathcal{E}_\theta(\rvx|\trvx) &=   f_\theta(\rvx) - \frac{1}{2\sigma^2} \|\trvx-\rvx\|^2 \\
        &\doteq f_\theta(\trvx) + \langle \nabla_\rvx f_\theta(\trvx), \rvx-\trvx\rangle  - \frac{1}{2\sigma^2} \|\trvx-\rvx\|^2\\  
        & = -\frac{1}{2\sigma^2} \left[ \|\rvx\|^2 - 2 \langle \trvx, \rvx \rangle + \|\trvx\|^2 \right] + \langle \nabla_\rvx f_\theta(\trvx), \rvx \rangle - \langle \nabla_\rvx f_\theta(\trvx), \trvx \rangle + f_\theta(\trvx) \\
        & = -\frac{1}{2\sigma^2} \left[ \|\rvx\|^2 - 2\langle\trvx + \sigma^2\nabla_\rvx f_\theta(\trvx), \rvx \rangle \right] - \frac{1}{2\sigma^2} \|\trvx\|^2 - \langle \nabla_\rvx f_\theta(\trvx), \trvx \rangle + f_\theta(\trvx) \\
        &= - \frac{1}{2\sigma^2} \left[ \|\rvx - (\trvx + \sigma^2 \nabla_\rvx f_\theta(\trvx))\|^2\right] + c,
\end{align}
 
\subsection{Difference between the scores of $p(\rvx)$ and $p(\trvx)$} \label{app:diff}
For notation clarity, with 
 $\trvx = \rvx + \beps$, we let $\widetilde{p}$ be the distribution of $\trvx$, and $p$ be the distribution of $\rvx$. Then for a smooth testing function with vanishing tails, 
 \begin{align} 
 \E[h(\trvx)] &= \E[h(\rvx + \beps)] \\
 & \doteq \E[h(\rvx) + h'(\rvx) \beps + h''(\rvx) \beps^2/2] \\
 &= \E[h(\rvx)] + \E[h''(\rvx)] \sigma^2/2. 
\end{align} 
Integral by parts, 
\begin{align} 
\E[h''(\rvx)] = \int h''(\rvx) p(\rvx) d\rvx = - \int h'(\rvx) p'(\rvx) d\rvx = \int p''(\rvx) h(\rvx) d\rvx. 
\end{align}
Thus we have the heat equation
\begin{eqnarray} 
\widetilde{p}(\rvx) = p(\rvx) + p''(\rvx) \sigma^2/2. 
\end{eqnarray} 
The score 
\begin{align} 
  \nabla_\rvx \log \tilde{p}(\rvx) &= \nabla_x \log p(\rvx) + \nabla_\rvx  \log (1 + p''(\rvx)/p(\rvx) \sigma^2/2) \\
  & \doteq \nabla_\rvx \log p(\rvx) + \nabla_\rvx [p''(\rvx)/p(\rvx)] \sigma^2/2. 
  \end{align} 
  Thus the difference between the score of $p$ and $\widetilde{p}$ is of the order $\sigma^2$, which is negligible when $\sigma^2$ is small. 

\subsection{Learning gradients of normal approximation and original recovery likelihood} \label{app:grad}
In this subsection we demonstrate that the learning gradient of maximizing likelihood of the normal approximation is approximately the same as the gradient of maximizing the original recovery likelihood with one step of Langevin sampling. Specifically, the gradient of the normal approximation of recovery log-likelihood for an observed $\rvx_{\rm obs}$ is
\begin{eqnarray} 
   \nabla_\theta \left(\frac{1}{2\sigma^2} \left[ \|\rvx_{\rm obs} - (\trvx + \sigma^2 f'_\theta(\trvx))\|^2\right]\right)  =  \nabla_\theta f'_\theta(\trvx)(\rvx_{\rm obs} - (\trvx + \sigma^2 f'_\theta(\trvx)). 
   \label{eq:grad1}
\end{eqnarray}
On the other hand, to maximize the original recovery likelihood, suppose we sample $\rvx_{\rm syn} \sim p_\theta(\rvx|\trvx)$, then the gradient ascent of the original recovery log-likelihood is
\begin{align} 
    \nabla_\theta f_\theta(\rvx_{\rm obs}) - \E[\nabla_\theta f_\theta(\rvx_{\rm syn})] = h_\theta(\rvx_{\rm obs}) - \E[h_\theta(\rvx_{\rm syn})], 
\end{align}
where $h_\theta(\rvx) = \nabla_\theta f_\theta(\rvx)$. Approximately, if we perform one step of Langevin dynamics from $\trvx$ to obtain $\rvx_{\rm syn}$, i.e., $x_{\rm syn} = \trvx + \sigma^2 f'_\theta(\trvx) + \sqrt{2}\sigma e$, and assume $f_\theta(\rvx)$ is locally linear in $\rvx$, then 
\begin{align} 
    & \nabla_\theta f_\theta(\rvx_{\rm obs}) - \E[\nabla_\theta f_\theta(\rvx_{\rm init})] \\
    &= 
          h_\theta(\rvx_{\rm obs}) - \E[h_\theta(\trvx + \sigma^2 f'_\theta(\trvx) + \sigma e)] \\
          & \doteq h_\theta(\trvx) + h'_\theta(\trvx) (x_{\rm obs} - \trvx) - 
          \E[h_\theta(\trvx) + h'_\theta(\trvx) (\sigma^2 f'_\theta(\trvx) + \sigma e)]\\
          &=h'_\theta(\trvx) (\rvx_{\rm obs} - (\trvx + \sigma^2 f'_\theta(\trvx)) \\
          &= \nabla_\theta f'_\theta(\trvx) (\rvx_{\rm obs} - (\trvx + \sigma^2 f'_\theta(\trvx)) .\label{eq:grad2}
\end{align}
Comparing equations \ref{eq:grad1} and \ref{eq:grad2}, we see that the two gradients agree with each other.

\subsection{Estimating the partition function} \label{app:ais}
We can utilize the sequence of learned distributions of $\rvy_t$ ($ = \sqrt{1 - \sigma_{t+1}^2} \rvx_t$) to estimate the partition function. Specifically, the marginal distribution of $\rvy_t$ is
\begin{align}
  p_\theta(\rvy_t) =  \frac{1}{Z_{\theta, t}} \exp\left(f_\theta(\rvy_t, t)\right)
\end{align}
We can estimate the ratio of the partition functions at two consecutive time steps using importance sampling
\begin{align}
\frac{Z_{\theta, t}}{Z_{\theta, t+1}} &= \E_{p_{\theta}(\rvy_{t+1})}\left[\exp(f_\theta(\rvy, t) - f_\theta(\rvy, t+1))\right] \\
 & \doteq \frac{1}{M} \sum_{i=1}^M\left[\exp(f_\theta(\rvy_{t+1, i}, t) - f_\theta(\rvy_{t+1, i}, t+1)) \right], \label{eq:ais}
\end{align}
where $\rvy_{t+1, i}$ are samples generated by progressive sampling.
Starting from $t= T$, where $p_T(x)$ follows Gaussian distribution, we can compute $\log Z_{\theta, t}$ along the reverse path of the diffusion process, until we reach $t=0$:
\begin{align}
     Z_{\theta, 0} = Z_{\theta, T} \prod_{t=0}^{T-1} \frac{Z_{\theta, t}}{Z_{\theta, t+1}}.
\end{align}
In practice, since the ratio given by MCMC samples can vary across many orders of magnitude, it is more meaningful to estimate \begin{eqnarray}
\log Z_{\theta, 0} = \log Z_{\theta, T} + \sum_{t=0}^{T-1} \log \frac{Z_{\theta, t}}{Z_{\theta, t+1}}. 
\end{eqnarray}
Unfortunately, although \eqref{eq:ais} is an unbiased estimator of $Z_{\theta, t} / Z_{\theta, t+1}$, the logarithm of this estimator is generally a stochastic lower bound of $\log(Z_{\theta, t} / Z_{\theta, t+1})$~\citep{grosse2016measuring}. However, as we show below, this bound will gradually converge to an unbiased estimator of $\log(Z_{\theta, t} / Z_{\theta, t+1})$, as the number of samples becomes large. Specifically, let $A$ be the estimator in \eqref{eq:ais}, $\mu$ be the true value of $Z_{\theta, t} / Z_{\theta, t+1}$. We have $\E[A] = \mu$, then by second order Taylor expansion,
\begin{align}
\E[\log A] &\doteq \E\left[\log \mu + \frac{1}{\mu} (A - \mu) - \frac{1}{2\mu^2}(A - \mu)^2\right]\\
&= \log \mu - \frac{1}{2\mu^2} \Var(A). 
\end{align}
By {\em law of large number}, $\Var(A) \to 0$ as $M \to \infty$, and thus $\E[\log A] \to \log \mu$. This is also consistent with the estimation curves in the right subfigure of \Figref{fig: long-run}: since $\Var(A) \geq 0$, the estimation curve increases from below as the number of samples becomes larger. When the curve becomes stable, it indicates the convergence. 

\section{Experimental details} \label{app:exp}
\paragraph{Model architecture.} Our network structure is based on Wide ResNet~\citep{zagoruyko2016wide}. Table \ref{tabl:structure} lists the detailed network structures of various resolutions. The number of ResBlocks at every level $N$ is a hyperparameter that we sweep over. The values of $N$ for various datasets are listed in Table \ref{tabl:hyper}. Each ResBlock consists of two Conv2D layers. For the second Conv2D layer, we use zero initialization for the weights, and add a trainable channel-wise scaling parameter to the output. We remove the weight normalization, and use leaky ReLU (slope $= 0.2$) as the activation function in ResBlocks. Spectral normalization~\citep{miyato2018spectral} is used to regularize parameters in Conv2D layer, ResBlocks and Dense layer. For encoding time step $t$, we follow the scheme in ~\citep{ho2020denoising}. Specifically, the time step $t$ is first transformed into sinusoidal embedding, and then two Dense layers is added. The time embedding is added after the first Conv2D layer of each ResBlock. 

\paragraph{Training.} We use Adam~\citep{kingma2014adam} optimizer for all the experiments. We find that for high resolution images, using a smaller $\beta_1$ in Adam help stabilize training. We use learning rate $0.0001$ for all the experiments. For the values of $\beta_1$, batch sizes and the number of training iterations for various datasets, see Table \ref{tabl:hyper}. 

\paragraph{Datasets.} We use the following datasets in our experiments: CIFAR-10~\citep{krizhevsky2009learning}, CelebA~\citep{liu2015faceattributes} and LSUN~\citep{yu2015lsun}. CIFAR-10 is of resolution $32 \times 32$, and contains $50,000$ training images and $10,000$ test images. CelebA contains 202,599 face images, of which 162,770 are training images and 19,962 are test images. For processing, we first clip each image to $178 \times 178$ and then resize it to $64 \times 64$. For LSUN, we use church\_outdoor and bedroom categories, which contains 126,227 and 3,033,042 training images respectively. Both categories contain $300$ test images. For processing, we first crop each image to a square image of the smaller size among the height and weight, and then we resize it to $64 \times 64$ or $128 \times 128$. For resizing, we set antialias to True. We apply horizontal random flip as data augmentation for all datasets during training. 

\paragraph{Evaluation metrics.} We use FID and inception scores as quantitative evaluation metrics of sample quality. On all the datasets, we calculate FID and inception scores on 50,000 samples using the original code from ~\citet{salimans2016improved} and ~\citet{heusel2017gans}. 

\begin{table}[ht]
\centering
 \caption{Model architectures of various solutions. $N$ is a hyperparameter that we sweep over.} 
 \footnotesize
 \begin{minipage}[t]{.3\textwidth}
 \centering
 (a) Resolution $32 \times 32$\\
 \begin{tabular}{c}
    \toprule
    $3 \times 3$ Conv2D, 128 \\
    \midrule
    $N$ ResBlocks, 128\\
    Downsample $2 \times 2$\\
    \midrule
    $N$ ResBlocks, 256 \\
    Downsample $2 \times 2$\\
    \midrule 
     $N$ ResBlocks, 256 \\
    Downsample $2 \times 2$\\
     \midrule 
     $N$ ResBlocks, 256 \\
    \midrule 
    ReLU, global sum\\
    Dense 1 \\
    \bottomrule 
\end{tabular}
 \end{minipage}
  \begin{minipage}[t]{.3\textwidth}
  \centering
 (b) Resolution $64 \times 64$ \\
 \begin{tabular}{c}
    \toprule
    $3 \times 3$ Conv2D, 128 \\
    \midrule
    $N$ ResBlocks, 128\\
    Downsample $2 \times 2$\\
    \midrule
    $N$ ResBlocks, 256 \\
    Downsample $2 \times 2$\\
    \midrule 
   $N$ ResBlocks, 256 \\
    Downsample $2 \times 2$\\
     \midrule 
   $N$ ResBlocks, 256 \\
    Downsample $2 \times 2$\\
     \midrule 
     $N$ ResBlocks, 512 \\
    \midrule 
    ReLU, global sum\\
    Dense 1 \\
    \bottomrule 
\end{tabular}
 \end{minipage}
  \begin{minipage}[t]{.3\textwidth}
  \centering
 (c) Resolution $128 \times 128$ \\
 \begin{tabular}{c}
    \toprule
    $3 \times 3$ Conv2D, 128 \\
    \midrule
    $N$ ResBlocks, 128\\
    Downsample $2 \times 2$\\
    \midrule
    $N$ ResBlocks, 256 \\
    Downsample $2 \times 2$\\
    \midrule 
   $N$ ResBlocks, 256 \\
    Downsample $2 \times 2$\\
     \midrule 
   $N$ ResBlocks, 256 \\
    Downsample $2 \times 2$\\
     \midrule 
    $N$ ResBlocks, 512 \\
    Downsample $2 \times 2$\\
     \midrule
     $N$ ResBlocks, 512 \\
    \midrule 
    ReLU, global sum\\
    Dense 1 \\
    \bottomrule 
\end{tabular}
 \end{minipage}\\
  \begin{minipage}[t]{.3\textwidth}
  \centering
 (d) Time embedding (temb) \\
 \begin{tabular}{c}
    \toprule
 sinusoidal embedding \\
 \midrule
 Dense, leakyReLU\\
 \midrule
Dense\\
    \bottomrule 
\end{tabular}
 \end{minipage} 
  \begin{minipage}[t]{.3\textwidth}
  \centering
 (e) ResBlock \\
 \begin{tabular}{c}
    \toprule
 leakyReLU, $3 \times 3$ Conv2D \\
 \midrule
 $+$ Dense(leakyReLU(temb))\\
 \midrule
 leakyReLU, $3 \times 3$ Conv2D\\
 \midrule
 + input \\
    \bottomrule 
\end{tabular}
 \end{minipage}
    \label{tabl:structure}
\end{table}

\begin{table}[ht]
    \centering
        \caption{Hyperparameters of various datasets.}
        \footnotesize
    \begin{tabular}{ccccc}
    \toprule
        Dataset & $N$ & $\beta_1$ in Adam & Batch size & Training iterations  \\
        \midrule
        CIFAR-10 & 8 & 0.9 & 256 & 240k \\
        CelebA & 6 & 0.5 & 128 & 880k \\
        LSUN church\_outdoor $64^2$ & 2 & 0.9 & 128 & 960k \\
        LSUN bedroom $64^2$ & 2 & 0.9 & 128 & 760k \\
        LSUN church\_outdoor $128^2$ & 2 & 0.5 & 64 & 840k \\
        LSUN bedroom $128^2$ & 5 & 0.5 & 64 & 580k \\
        \bottomrule
    \end{tabular}
    \label{tabl:hyper}
\end{table}
\section{Additional experimental results}
\subsection{FID scores over iterations} \label{app:fid}
\begin{figure}[h]
 \centering
 \includegraphics[width=.5\textwidth]{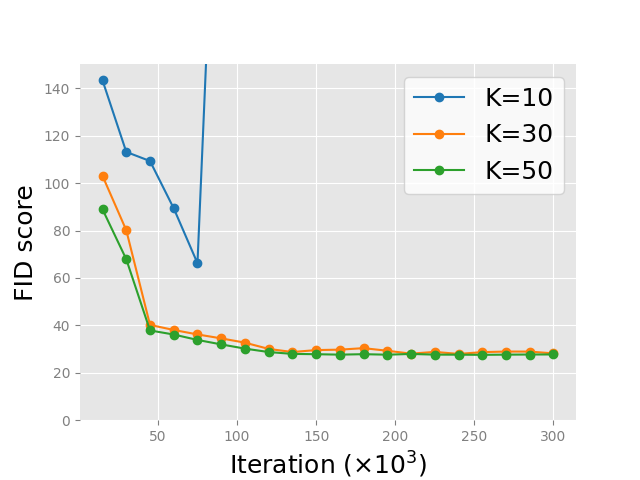} 
 \caption{FIDs for different number of Langevin steps.}
 \label{fig:fid}
 \end{figure}

Figure \ref{fig:fid} demonstrates FID scores computed on 2,500 samples every 15,000 iterations. 
\subsection{Long-run Chain sampling with NUTS} \label{app:long-run}
\begin{figure}[ht]
\begin{center}
\begin{tabular}{ccc}
	\includegraphics[width=.3\textwidth]{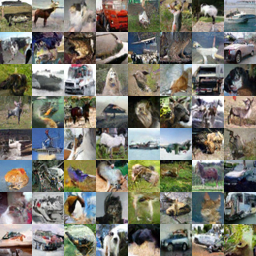} &
\includegraphics[width=.3\textwidth]{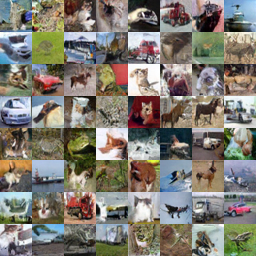} &
\includegraphics[width=.3\textwidth]{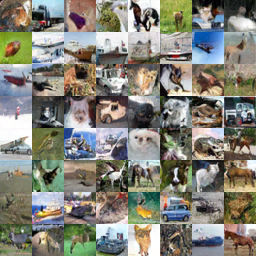}\\
(a) $1k$ steps, FID=24.78 & (b) $10k$ steps, FID=23.89 & (c) $100k$ steps, FID=25.08
\end{tabular}
\end{center}
\caption{Long run chain samples with different total number of NUTS steps. }
\label{fig: long-run2}
\end{figure}

As a further check, we use a No-U-Turn Sampler~\citep{hoffman2014no} to perform the long-run chain sampling, with the same step size schedule obtained for HMC sampler. Figure \ref{fig: long-run2} displays samples with different number of sampling steps. The samples remain realistic after $100k$ sampling steps in total and the FID score remains stable. 

\subsection{Additional interpolation results} \label{app:interp}
\begin{figure}[ht]
\begin{center}
\includegraphics[width=\textwidth]{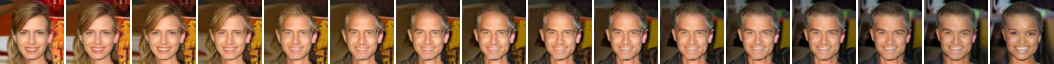}
\includegraphics[width=\textwidth]{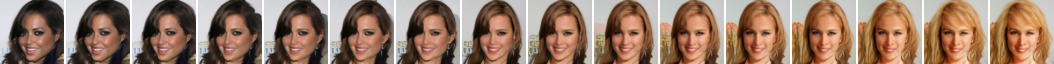}
\includegraphics[width=\textwidth]{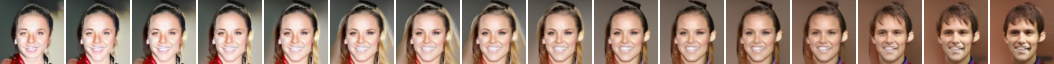}
\includegraphics[width=\textwidth]{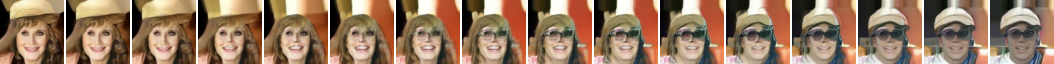}
\includegraphics[width=\textwidth]{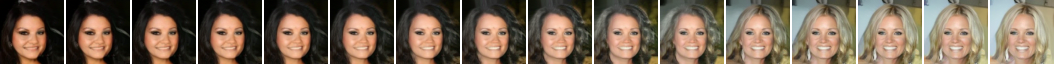}
\includegraphics[width=\textwidth]{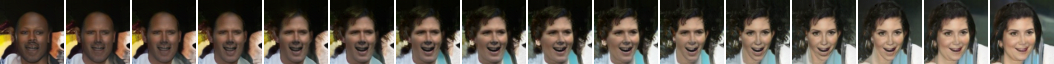}
\end{center}
\caption{Interpolation results between the leftmost and rightmost generated samples on CelebA $64 \times 64$. }
\label{fig: interp-celeba}
\end{figure}

\begin{figure}[ht]
\begin{center}
\includegraphics[width=\textwidth]{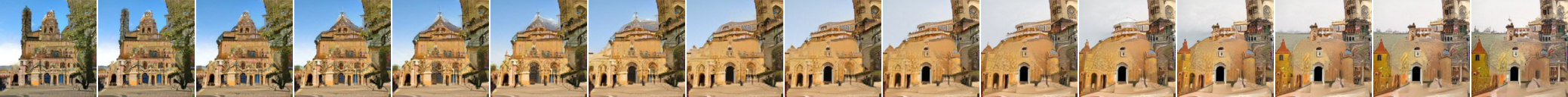}
\includegraphics[width=\textwidth]{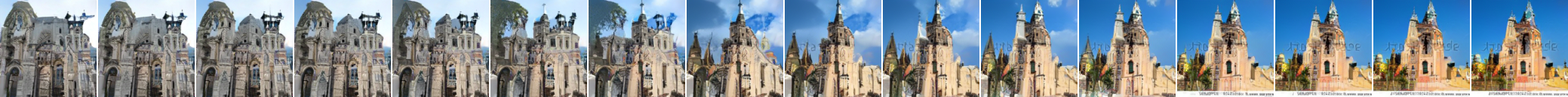}
\includegraphics[width=\textwidth]{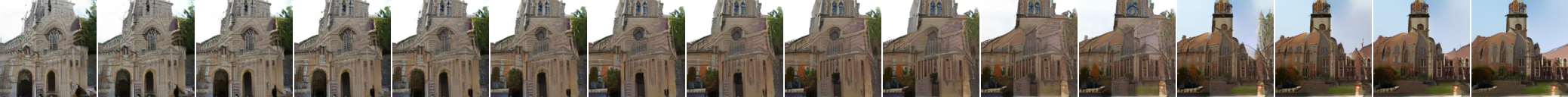}
\includegraphics[width=\textwidth]{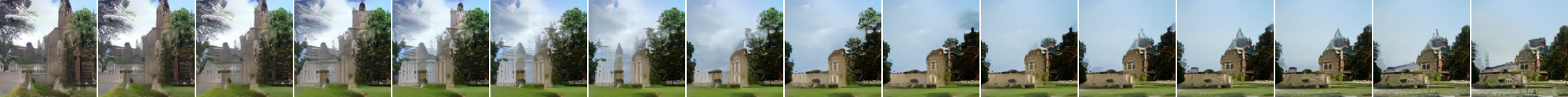}
\includegraphics[width=\textwidth]{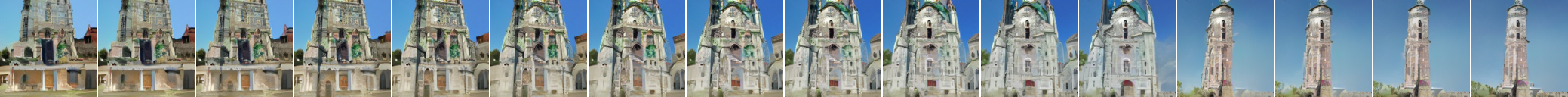}
\includegraphics[width=\textwidth]{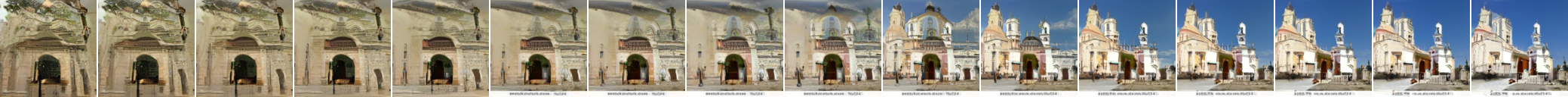}
\end{center}
\caption{Interpolation results between the leftmost and rightmost generated samples on LSUN church\_outdoor $128 \times 128$. }
\label{fig: interp-church}
\end{figure}

\begin{figure}[ht]
\begin{center}
\includegraphics[width=\textwidth]{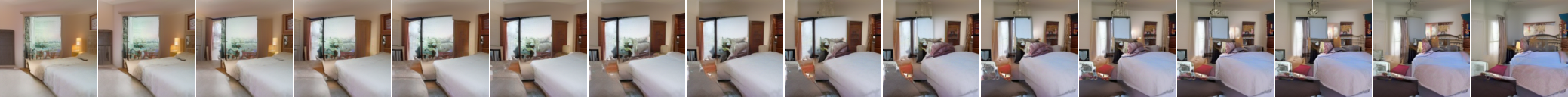}
\includegraphics[width=\textwidth]{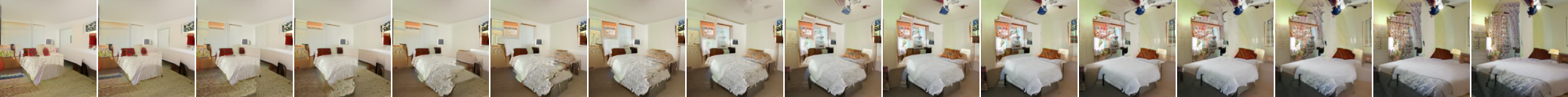}
\includegraphics[width=\textwidth]{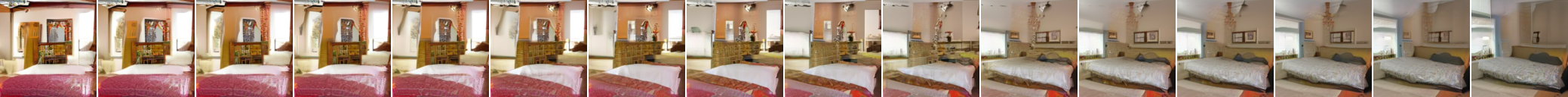}
\includegraphics[width=\textwidth]{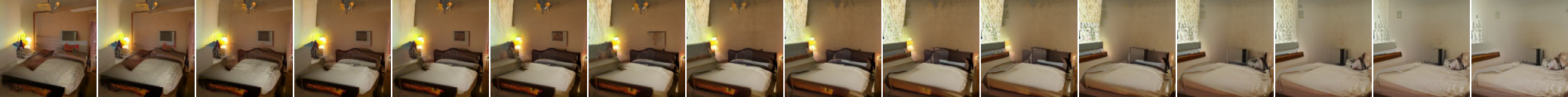}
\includegraphics[width=\textwidth]{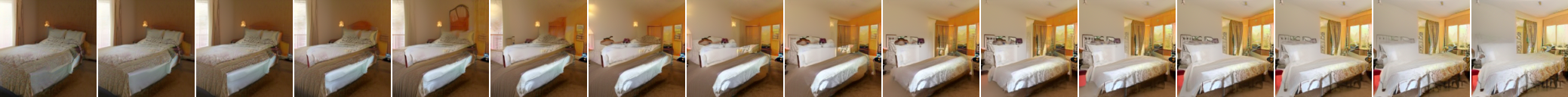}
\includegraphics[width=\textwidth]{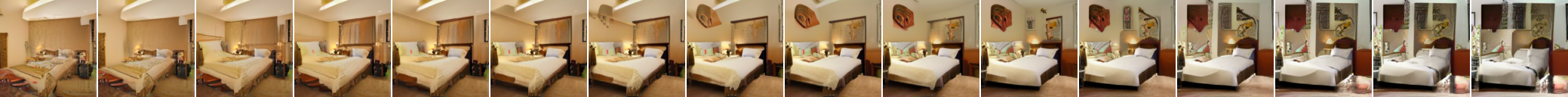}
\end{center}
\caption{Interpolation results between the leftmost and rightmost generated samples on LSUN bedroom $128 \times 128$. }
\label{fig: interp-bedroom}
\end{figure}

Figures \ref{fig: interp-celeba}, \ref{fig: interp-church} and \ref{fig: interp-bedroom} display more examples of interpolation between two generated samples on CelebA $64^2$, LSUN church\_outdoor $128^2$ and LSUN bedroom $128^2$.

\subsection{Additional image inpainting results} \label{app:inpaint}
\begin{figure}[ht]
\begin{center}
\includegraphics[width=.8\textwidth]{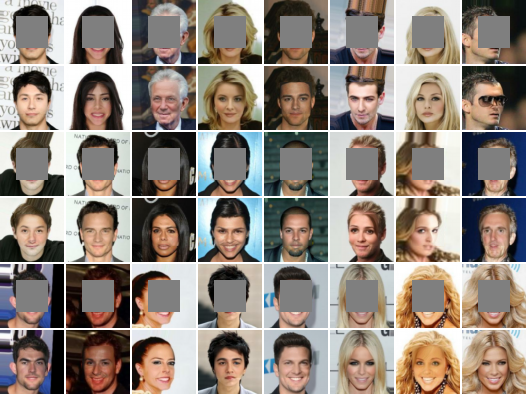} 
\end{center}
\caption{Image inpainting results on CelebA $64 \times 64$. {\em Top}: masked images, {\em bottom}: inpainted images.}
\label{fig: inpaint-celeba}
\end{figure}

\begin{figure}[ht]
\begin{center}
\includegraphics[width=\textwidth]{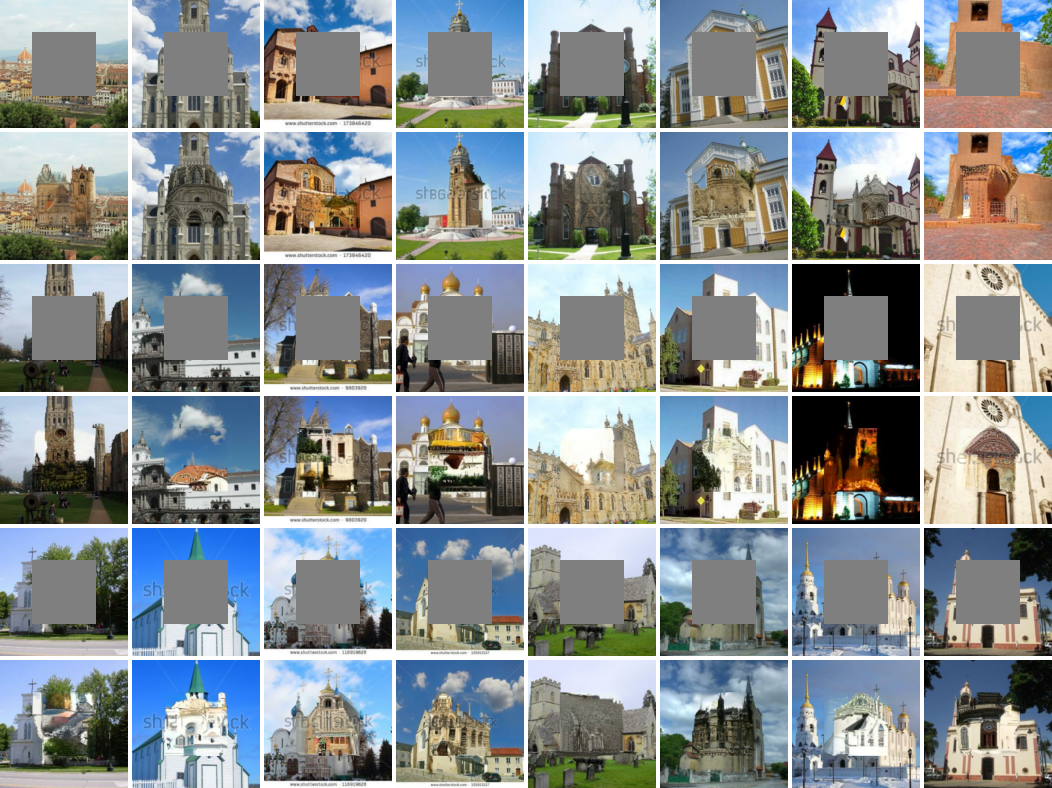} 
\end{center}
\caption{Image inpainting results on LSUN church\_outdoor $128 \times 128$. {\em Top}: masked images, {\em bottom}: inpainted images.}
\label{fig: inpaint-church}
\end{figure}

Figures \ref{fig: inpaint-celeba} and \ref{fig: inpaint-church} show additional examples of image inpainting on CelebA $64^2$ and LSUN church\_outdoor $128^2$. 

\subsection{Additional uncurated samples} \label{app:samples}
\begin{figure}[ht]
\begin{center}
\includegraphics[width=\textwidth]{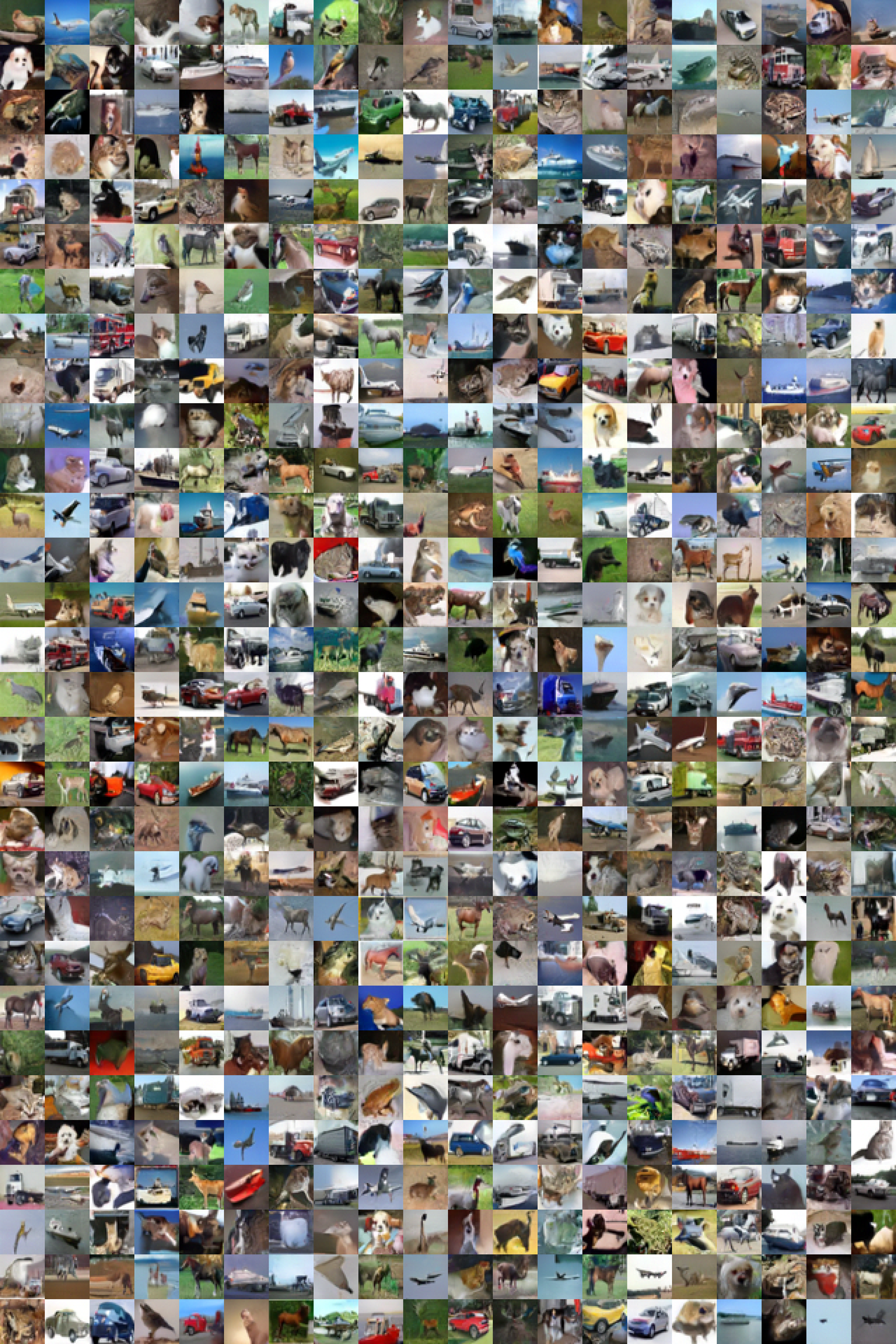}
\end{center}
\caption{Generated samples on CIFAR-10. }
\label{fig:cifar}
\end{figure}

\begin{figure}[ht]
\begin{center}
\includegraphics[width=\textwidth]{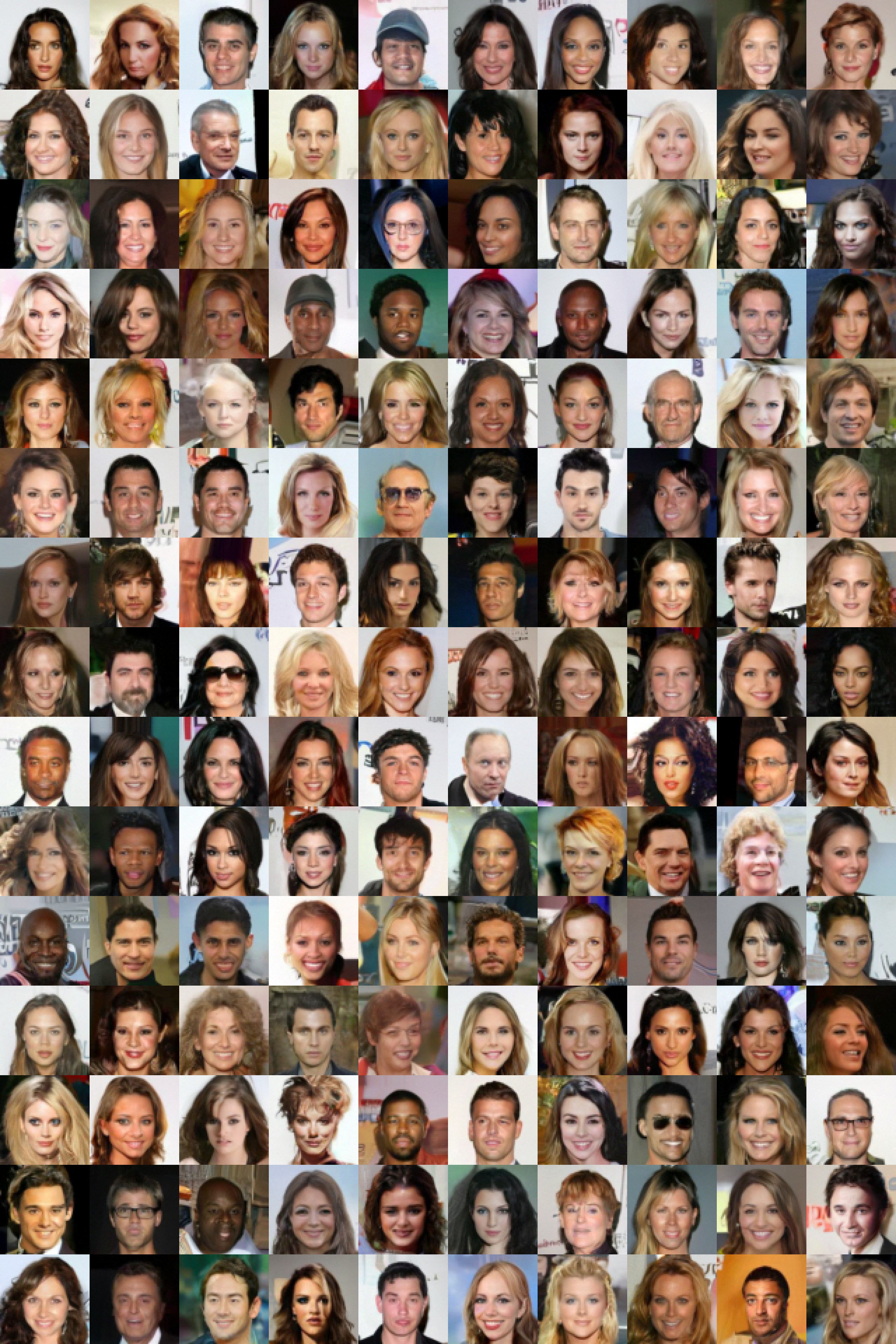}
\end{center}
\caption{Generated samples on CelebA $64 \times 64$. }
\label{fig:celeba}
\end{figure}

\begin{figure}[ht]
\begin{center}
\includegraphics[width=\textwidth]{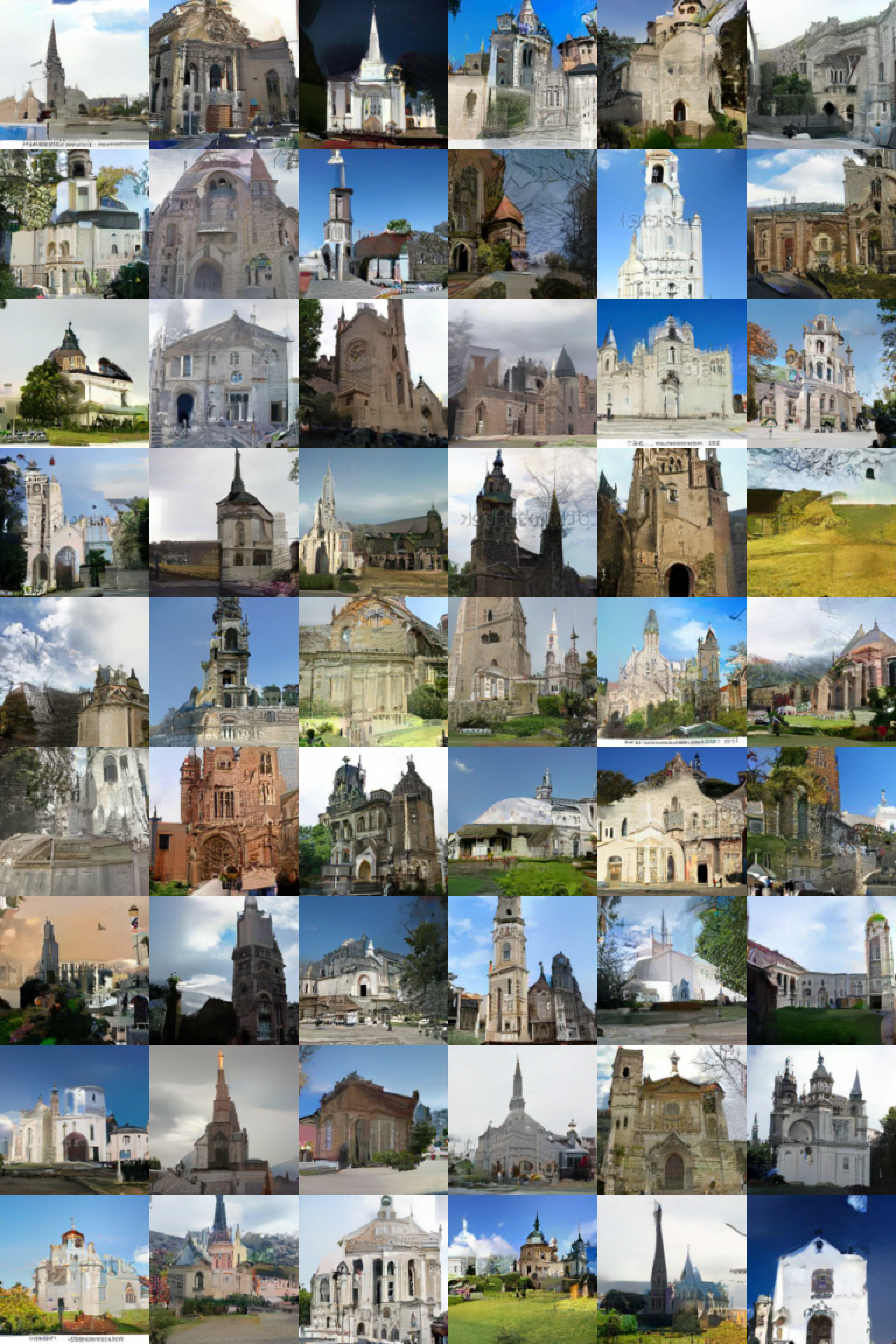}
\end{center}
\caption{Generated samples on LSUN church\_outdoor $128 \times 128$. FID=9.76}
\label{fig:church_128}
\end{figure}

\begin{figure}[ht]
\begin{center}
\includegraphics[width=\textwidth]{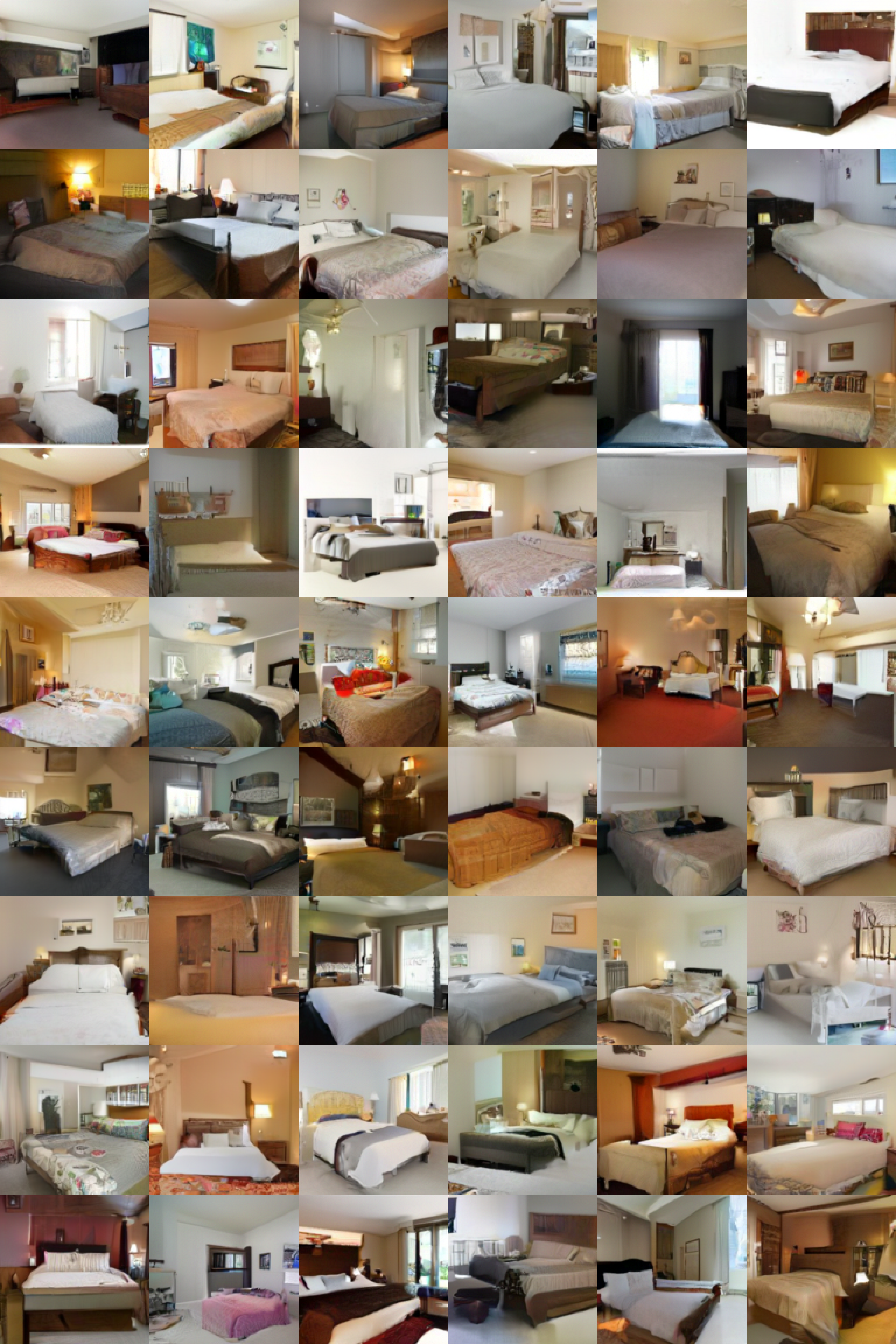}
\end{center}
\caption{Generated samples on LSUN bedroom $128 \times 128$. FID=11.27}
\label{fig:bedroom_128}
\end{figure}

\begin{figure}[ht]
\begin{center}
\includegraphics[width=\textwidth]{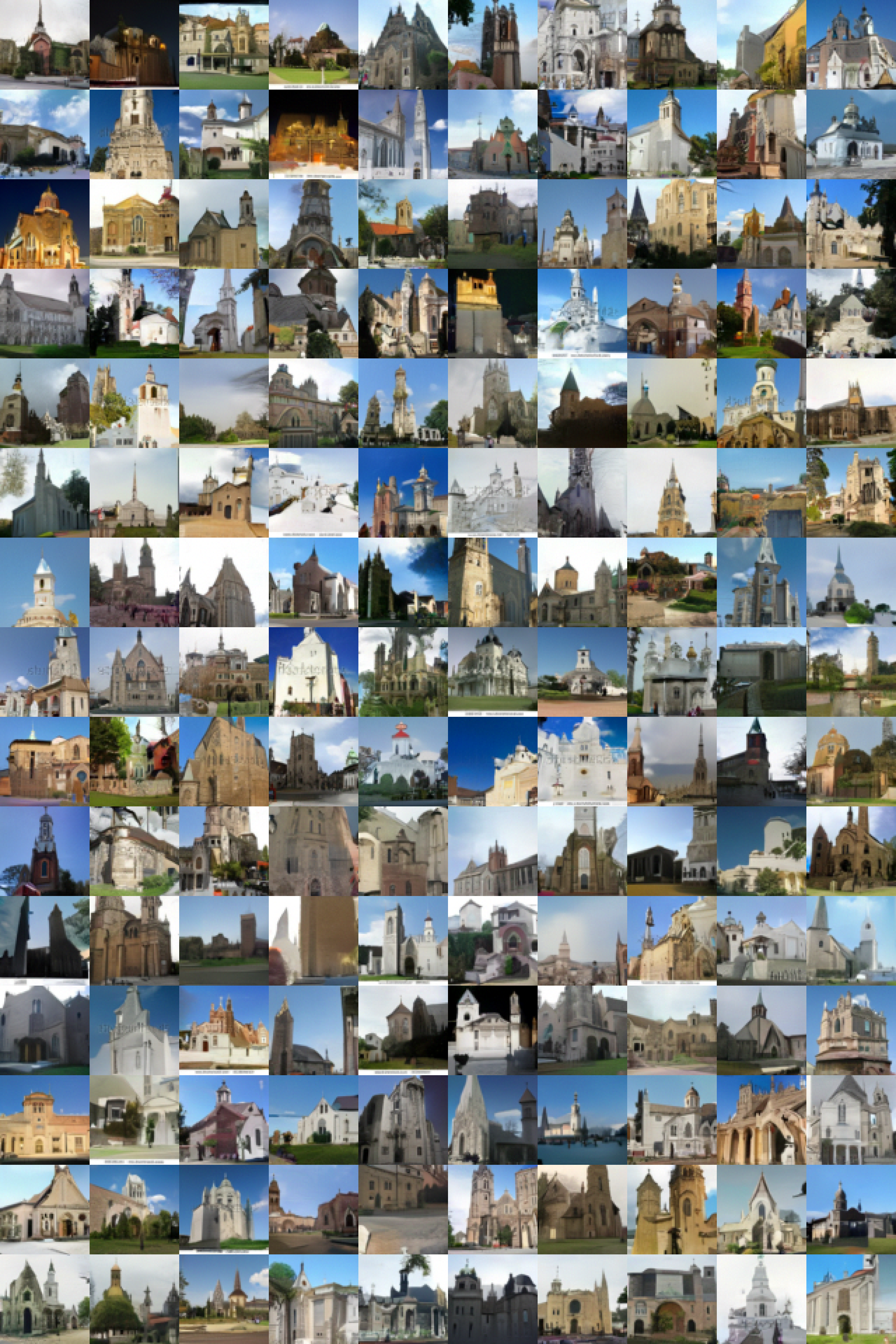}
\end{center}
\caption{Generated samples on LSUN church\_outdoor $64 \times 64$. FID=7.02}
\label{fig:church_64}
\end{figure}

\begin{figure}[ht]
\begin{center}
\includegraphics[width=\textwidth]{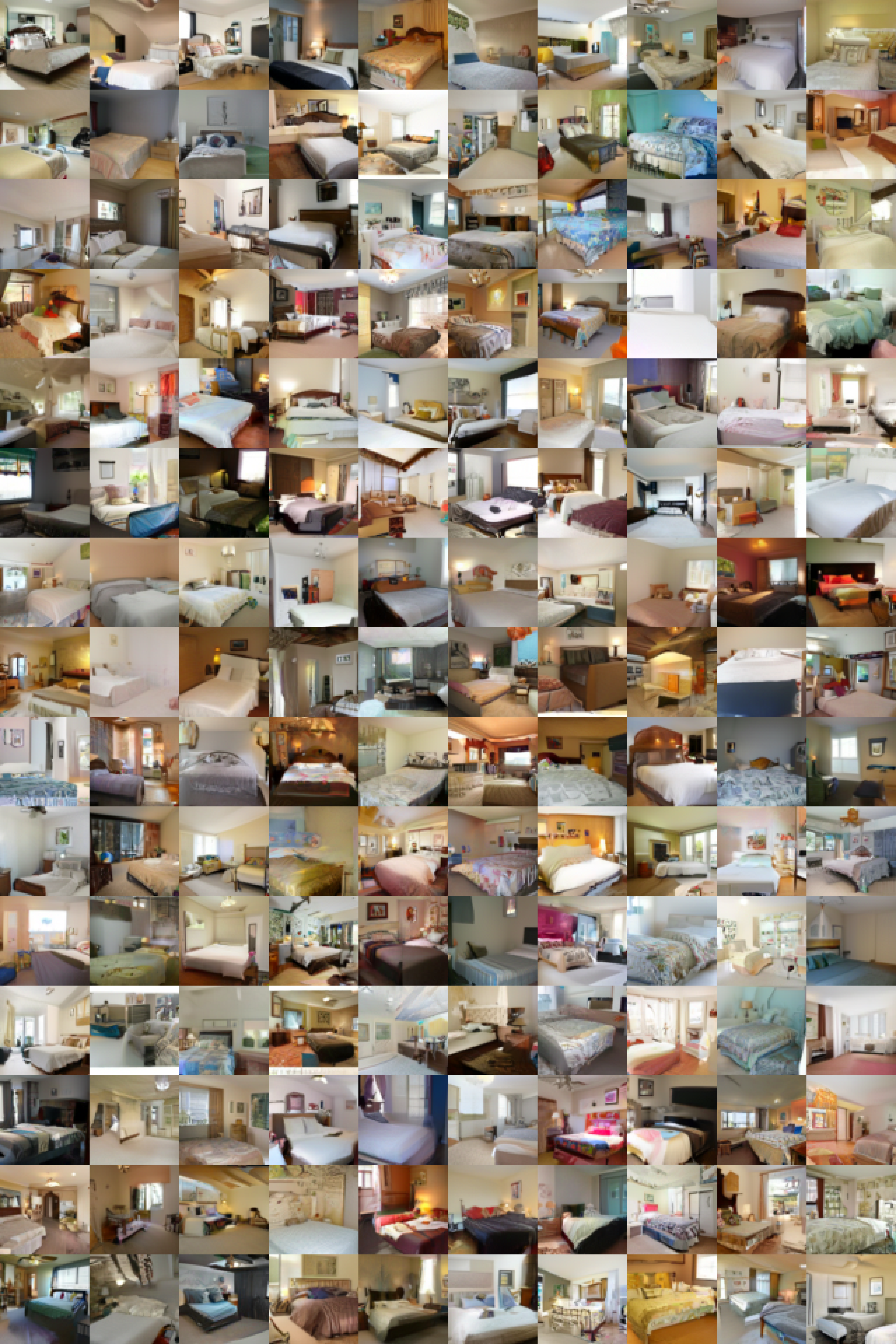}
\end{center}
\caption{Generated samples on LSUN bedroom $64 \times 64$. FID=8.98}
\label{fig:bedroom_64}
\end{figure}

Figures \ref{fig:cifar}, \ref{fig:celeba}, \ref{fig:church_128}, \ref{fig:bedroom_128}, \ref{fig:church_64} and \ref{fig:bedroom_64} show uncurated samples from the learned models under {\em T6} setting on CIFAR-10, CelebA $64^2$, LSUN church\_outdoor $128^2$, LSUN bedroom $128^2$, LSUN church\_outdoor $64^2$ and LSUN bedroom $64^2$ datasets.

\end{document}